\documentclass[sigconf,nonacm]{acmart}
\usepackage{amsmath}
\usepackage{tabularx}
\usepackage{float}
\usepackage{xcolor}
\usepackage{needspace}
\newcommand{\RT}[1]{Experiment~#1}
\newcommand{\BenchmarkEightFinal}{26.7\%}
\newcommand{\BenchmarkFinal}{28.2\%}

\newcommand{\BridgeGptOperating}{4.8}

\newcommand{\BridgeGptStaking}{0.6}
\newcommand{\BridgeGptTotal}{18.1}
\newcommand{\BridgeKimiModels}{3.0}
\newcommand{\BridgeKimiOperating}{3.6}

\newcommand{\BridgeKimiStaking}{2.5}
\newcommand{\BridgeKimiTotal}{9.1}

\newcommand{\CrossLgbmBench}{0.63}
\newcommand{\CrossLgbmFit}{0.023}
\newcommand{\CrossLgbmIc}{0.115}

\newcommand{\CrossLgbmInov}{0.018}
\newcommand{\CrossLgbmLadFit}{0.030}
\newcommand{\CrossLgbmLadP}{$p=0.002$}

\newcommand{\CrossLgbmOneFit}{0.024}
\newcommand{\CrossLgbmOneN}{16}
\newcommand{\CrossLgbmRet}{17.6\%}

\newcommand{\CrossLgbmUnq}{0.001}
\newcommand{\CrossNovShare}{69\%}

\newcommand{\CrossScoreSdRatio}{3.5}
\newcommand{\CrossSewBench}{0.55}
\newcommand{\CrossSewFit}{0.027}
\newcommand{\CrossSewIc}{0.108}

\newcommand{\CrossSewInov}{0.022}
\newcommand{\CrossSewLadFit}{0.027}

\newcommand{\CrossSewRet}{22.7\%}

\newcommand{\CrossSewUnq}{0.006}
\newcommand{\CrossTermP}{$p<0.001$}

\newcommand{\FableEightMean}{7.3\%}

\newcommand{\FableOpusCostRatio}{10}
\newcommand{\FlipAgents}{8}

\newcommand{\FlipHoldoutAgents}{5}

\newcommand{\FlipStakes}{25}
\newcommand{\FlipStakesLost}{24}
\newcommand{\FlipStakesP}{$p<0.001$}

\newcommand{\GptLive}{7.3\%}

\newcommand{\GptNever}{8}
\newcommand{\GptNeverCI}{[28\%, 72\%]}
\newcommand{\GptPooled}{2.2\%}
\newcommand{\GptPooledN}{16}

\newcommand{\HaikuEightN}{8}

\newcommand{\HaikuEightSubmittedNone}{7}

\newcommand{\KimiNever}{2}

\newcommand{\KimiPooled}{11.1\%}
\newcommand{\KimiPooledN}{16}
\newcommand{\KimiStakeEffect}{$-$2.2}
\newcommand{\LabelLag}{21}
\newcommand{\LadAllFeat}{139}
\newcommand{\LadAllFeatOf}{157}
\newcommand{\LadAllRoundsPos}{98\%}
\newcommand{\LadBetweenSimMax}{0.98}
\newcommand{\LadBetweenSimMin}{0.35}
\newcommand{\LadClaimBigAll}{39}
\newcommand{\LadClaimN}{162}

\newcommand{\LadEightModels}{4}

\newcommand{\LadEnsBetter}{37}
\newcommand{\LadEnsCells}{39}

\newcommand{\LadExcluded}{11}

\newcommand{\LadFableBest}{0.135}
\newcommand{\LadFableBestRung}{8}

\newcommand{\LadFableFree}{0.134}

\newcommand{\LadFableOne}{0.109}
\newcommand{\LadFableOpusCost}{3.7}

\newcommand{\LadFableRandom}{0}

\newcommand{\LadFableSonnetCost}{15}

\newcommand{\LadGbm}{99}
\newcommand{\LadGbmOf}{100}

\newcommand{\LadHaikuClaim}{+0.243}
\newcommand{\LadHaikuClaimBig}{16}

\newcommand{\LadHaikuCounted}{50}
\newcommand{\LadHaikuFree}{0.108}
\newcommand{\LadHaikuIds}{15}

\newcommand{\LadHaikuOne}{0.105}

\newcommand{\LadOneBetween}{0.009}
\newcommand{\LadOneExplained}{18\%}

\newcommand{\LadOneFeatures}{22}

\newcommand{\LadOneMedianHi}{0.115}
\newcommand{\LadOneMedianLo}{0.105}
\newcommand{\LadOneModal}{4}
\newcommand{\LadOneModalOf}{20}

\newcommand{\LadOneP}{$p=0.34$}

\newcommand{\LadOneWithin}{0.019}
\newcommand{\LadOpusBest}{0.133}
\newcommand{\LadOpusBestRung}{5}

\newcommand{\LadOpusClaimBig}{5}

\newcommand{\LadOpusFree}{0.124}

\newcommand{\LadOpusOne}{0.108}

\newcommand{\LadOpusRandom}{5}

\newcommand{\LadOpusSimMax}{0.98}
\newcommand{\LadOpusSimMin}{0.65}

\newcommand{\LadOrderedBig}{3}
\newcommand{\LadOrderedClaim}{$-$0.009}
\newcommand{\LadOrderedN}{115}
\newcommand{\LadOrderedTunedMax}{+0.037}
\newcommand{\LadOrderedTunedN}{59}
\newcommand{\LadOtherInvalid}{one other invalid run}
\newcommand{\LadOtherInvalidWhat}{was still training when the 45-minute cap ended it and wrote no predictions (Opus 5.5 on rung~10)}
\newcommand{\LadPracticeNeg}{96\%}
\newcommand{\LadRandomBig}{26}
\newcommand{\LadRandomN}{28}

\newcommand{\LadRungs}{10}
\newcommand{\LadRungsLuck}{7}

\newcommand{\LadRungsSplitList}{3, 5, 8 and 10}
\newcommand{\LadRuns}{200}

\newcommand{\LadRunsMid}{5}

\newcommand{\LadSevenAux}{12}

\newcommand{\LadSevenEns}{19}

\newcommand{\LadSevenModels}{3}
\newcommand{\LadSevenN}{20}

\newcommand{\LadSevenWithin}{0.008}
\newcommand{\LadSixAux}{1}

\newcommand{\LadSixN}{20}

\newcommand{\LadSonnetClaimBig}{18}

\newcommand{\LadSonnetFree}{0.116}

\newcommand{\LadSonnetOne}{0.115}

\newcommand{\LadSonnetRandom}{18}
\newcommand{\LadSonnetRandomOf}{35}
\newcommand{\LadSonnetSimMax}{0.92}
\newcommand{\LadSonnetSimMin}{0.56}

\newcommand{\LadTenWithin}{0.028}

\newcommand{\LadThreeExplained}{71\%}
\newcommand{\LadThreeFable}{0.107}
\newcommand{\LadThreeFableClaim}{+0.009}
\newcommand{\LadThreeHaiku}{0.061}
\newcommand{\LadThreeHaikuClaim}{+0.448}
\newcommand{\LadThreeHolm}{$p=0.07$}
\newcommand{\LadThreeOpus}{0.115}
\newcommand{\LadThreeOpusClaim}{$-$0.004}
\newcommand{\LadThreeP}{$p=0.004$}
\newcommand{\LadThreeSonnet}{0.056}
\newcommand{\LadThreeSonnetClaim}{+0.400}

\newcommand{\LadTwoClaimMax}{+0.049}
\newcommand{\LadTwoExplained}{23\%}

\newcommand{\LadTwoOpus}{0.115}

\newcommand{\LadTwoP}{$p=0.17$}
\newcommand{\LadTwoSonnet}{0.116}

\newcommand{\LadValid}{184}
\newcommand{\LadWorstClaim}{0.87}
\newcommand{\LadWorstIc}{0.017}

\newcommand{\LadWorstRung}{8}
\newcommand{\LgbmSd}{5.4}
\newcommand{\LiveAgents}{91}
\newcommand{\LiveBenchFirst}{0.228}
\newcommand{\LiveBenchLast}{0.097}
\newcommand{\LiveCfN}{7}
\newcommand{\LiveCfWorstActual}{5.2\%}
\newcommand{\LiveCfWorstAlt}{80.0\%}
\newcommand{\LiveCfWorstWhat}{same models, not flipped in round 1}
\newcommand{\LiveConduct}{8}
\newcommand{\LiveConversations}{296}
\newcommand{\LiveEvents}{7}
\newcommand{\LiveFinishedN}{24}

\newcommand{\LiveFlippedHi}{$-$10.1}
\newcommand{\LiveFlippedLo}{$-$34.3}
\newcommand{\LiveFlippedN}{5}
\newcommand{\LiveIssues}{14}

\newcommand{\LiveMedianFirst}{0.101}
\newcommand{\LiveMedianLast}{0.055}
\newcommand{\LiveModelsMedian}{8}
\newcommand{\LiveModelsRho}{$-$0.09}
\newcommand{\LiveModelsRhoP}{$p=0.69$}
\newcommand{\LiveNormalHi}{21.8}
\newcommand{\LiveNormalLo}{11.9}
\newcommand{\LiveNormalN}{6}

\newcommand{\LiveNycAgents}{49}
\newcommand{\LiveNycAnalysed}{36}
\newcommand{\LiveNycBench}{94.6\%}
\newcommand{\LiveNycBest}{75.6\%}

\newcommand{\LiveNycHosted}{40}

\newcommand{\LiveNycLead}{Opus 5.5 (16), Fable 5.1 (6), Kimi K3 (5), Sonnet 5 (4), GPT-5.6 Sol (3), GPT-6 Astra (2)}
\newcommand{\LiveNycMedian}{33.0\%}
\newcommand{\LiveNycNoResult}{14}

\newcommand{\LivePracExampleN}{6}

\newcommand{\LivePracOneHist}{0.46}
\newcommand{\LivePracOnePrac}{0.00}
\newcommand{\LivePracTwoOne}{0.78}
\newcommand{\LivePracTwoPrac}{$-$0.11}

\newcommand{\LiveTokMedian}{39}
\newcommand{\LiveTokRho}{$-$0.31}
\newcommand{\LiveTokRhoP}{$p=0.14$}
\newcommand{\LossFableN}{8}
\newcommand{\LossFableRank}{6}
\newcommand{\LossFreeOpusN}{8}
\newcommand{\LossGptN}{8}

\newcommand{\LossHaikuN}{7}

\newcommand{\LossKimiN}{8}

\newcommand{\LossKimiReg}{6}
\newcommand{\LossOpusN}{48}

\newcommand{\LossSonnetN}{8}
\newcommand{\LossSonnetRank}{7}

\newcommand{\NPerGroupBonf}{27}
\newcommand{\NPerGroupOneSd}{17}

\newcommand{\NeffOpusSeven}{1.2}

\newcommand{\OpusEightLiveMean}{13.5\%}
\newcommand{\OpusEightLiveN}{6}
\newcommand{\OpusEightMean}{10.1\%}
\newcommand{\OpusEightN}{8}
\newcommand{\OpusEightNever}{2}

\newcommand{\OpusFullEqual}{19.9\%}

\newcommand{\OpusMinusFableEight}{2.8}
\newcommand{\OpusMinusFableEightCI}{$[-2.5, 7.9]$}
\newcommand{\OpusMinusFableEightP}{$p=0.34$}
\newcommand{\OpusMinusGpt}{18.1}
\newcommand{\OpusMinusGptCI}{$[13.7, 22.2]$}
\newcommand{\OpusMinusGptP}{$p<0.001$}
\newcommand{\OpusMinusKimi}{9.1}
\newcommand{\OpusMinusKimiCI}{$[3.3, 15.2]$}
\newcommand{\OpusMinusKimiHolm}{$p=0.11$}
\newcommand{\OpusMinusKimiP}{$p=0.006$}
\newcommand{\OpusMinusSonnetEight}{1.2}
\newcommand{\OpusMinusSonnetEightCI}{$[-5.5, 7.4]$}
\newcommand{\OpusMinusSonnetEightP}{$p=0.73$}

\newcommand{\OpusPooled}{20.2\%}
\newcommand{\OpusPooledN}{16}
\newcommand{\OpusStakeEffect}{0.3}
\newcommand{\OtherStakes}{714}
\newcommand{\OtherStakesLost}{16}
\newcommand{\OwnAllIn}{11}

\newcommand{\OwnMinusToldCI}{$[-3.2, 3.2]$}
\newcommand{\OwnMinusToldP}{$p=0.98$}
\newcommand{\OwnN}{16}
\newcommand{\OwnShareLater}{99\%}
\newcommand{\OwnShareRoundOne}{85\%}

\newcommand{\PFamilyN}{27}
\newcommand{\PHolmSig}{7}
\newcommand{\PRawSig}{14}

\newcommand{\PracticeModels}{361}

\newcommand{\PracticeNegHi}{78\%}
\newcommand{\PracticeNegLo}{58\%}

\newcommand{\PracticeRho}{$-$0.62}
\newcommand{\PracticeRhoP}{$p<0.001$}
\newcommand{\PracticeRhoUnflipped}{$-$0.61}

\newcommand{\SewMinusLgbm}{5.1}
\newcommand{\SewMinusLgbmCI}{$[2.5, 7.7]$}
\newcommand{\SewMinusLgbmP}{$p=0.002$}
\newcommand{\SewSd}{1.1}
\newcommand{\SimKimiMax}{0.51}
\newcommand{\SimKimiMin}{0.44}
\newcommand{\SimOpusFable}{0.53}
\newcommand{\SimOpusGptMax}{0.56}
\newcommand{\SimOpusGptMin}{0.16}
\newcommand{\SimOpusKimiMax}{0.65}
\newcommand{\SimOpusKimiMin}{0.61}
\newcommand{\SimOpusMax}{0.98}
\newcommand{\SimOpusMin}{0.81}
\newcommand{\SimOpusSeven}{0.81}
\newcommand{\SimOpusSonnet}{0.84}

\newcommand{\SonnetEightMean}{8.9\%}

\newcommand{\SplitBoardGap}{88}
\newcommand{\SplitTrainGap}{41}
\newcommand{\SpreadP}{$p=0.002$}

\newcommand{\SpreadShare}{0.2}

\newcommand{\TokHighN}{118}
\newcommand{\TokHighNever}{10}
\newcommand{\TokLowGpt}{5}
\newcommand{\TokLowHaiku}{3}
\newcommand{\TokLowKimi}{2}
\newcommand{\TokLowN}{10}
\newcommand{\TokOverTen}{13 and 18}
\newcommand{\TokRho}{+0.08}
\newcommand{\TokRhoP}{$p=0.45$}
\newcommand{\TokStakedRho}{-0.20}
\newcommand{\TokStakedRhoP}{$p=0.06$}
\newcommand{\TokTenAtTen}{59}
\newcommand{\TokTenN}{92}
\newcommand{\TokTenRho}{+0.21}
\newcommand{\TokTenRhoOpus}{+0.32}
\newcommand{\TokTenRhoOpusP}{$p=0.04$}
\newcommand{\TokTenRhoP}{$p=0.05$}

\newcommand{\EightTable}{%
\begin{tabular}{@{}lrrcrr@{}}
\toprule
\begin{tabular}[b]{@{}l@{}}Language\\model\end{tabular} & $n$ & Return (\%) & 95\% CI & Never staked & Cost (\$) \\
\midrule
Opus 5.5 & 8 & 10.1 & $[5.0, 14.8]$ & 2 & 2.12 \\
Sonnet 5 & 8 & 8.9 & $[5.1, 13.4]$ & 0 & 6.26 \\
Fable 5.1 & 8 & 7.3 & $[5.3, 9.1]$ & 0 & 20.75 \\
Haiku 4.5 & 8 & 0.0 & $[0.0, 0.0]$ & 8 & 0.92 \\
\bottomrule
\end{tabular}}
\newcommand{\PromptTable}{%
\begin{tabular}{@{}lllrrcr@{}}
\toprule
Test & Model asked for & Prompt & $n$ & Return (\%) & 95\% CI & SD (points) \\
\midrule
Experiment 1 & Signed equal-weight & 1 model, all in & 7 & 23.3 & $[22.8, 24.0]$ & 0.9 \\
Experiment 1 & Signed equal-weight & up to 10, chosen models & 8 & 22.1 & $[21.5, 22.7]$ & 0.9 \\
Experiment 2 & LightGBM on ranks & 1 model, all in & 8 & 17.7 & $[13.9, 21.3]$ & 5.8 \\
Experiment 2 & LightGBM on ranks & up to 10, chosen models & 8 & 17.5 & $[14.2, 20.9]$ & 5.3 \\
Experiment 3 & LightGBM on ranks & 1 model, own staking & 8 & 17.8 & $[15.2, 20.6]$ & 4.2 \\
Experiment 3 & LightGBM on ranks & up to 10, own staking & 8 & 17.5 & $[14.9, 19.9]$ & 3.9 \\
\bottomrule
\end{tabular}}
\newcommand{\LlmTable}{%
\begin{tabular}{@{}llrrcrr@{}}
\toprule
Test & Language model & $n$ & Return (\%) & 95\% CI & Range (\%) & Never staked \\
\midrule
Experiment 3 & Opus 5.5 & 8 & 20.9 & $[17.1, 24.1]$ & 9.6 to 26.7 & 0 \\
Experiment 3 & Kimi K3 & 8 & 13.1 & $[3.3, 20.7]$ & $-$15.7 to 25.0 & 0 \\
Experiment 3 & GPT-5.6 Sol & 8 & 2.5 & $[-3.5, 8.7]$ & $-$10.3 to 17.1 & 2 \\
\addlinespace
Experiment 4 & Opus 5.5 & 8 & 19.6 & $[15.2, 23.3]$ & 6.2 to 28.1 & 0 \\
Experiment 4 & Kimi K3 & 8 & 9.2 & $[3.8, 15.0]$ & 0.0 to 23.6 & 2 \\
Experiment 4 & GPT-5.6 Sol & 8 & 1.9 & $[0.0, 4.3]$ & 0.0 to 8.3 & 6 \\
\bottomrule
\end{tabular}}
\newcommand{\SimTable}{%
\begin{tabular}{@{}llrr@{}}
\toprule
Test & Language model & Pairs & Mean similarity \\
\midrule
Experiment 1 & Opus 5.5 & 56 & 0.98 \\
Experiment 2 & Opus 5.5 & 56 & 0.87 \\
Experiment 3 & Opus 5.5 & 60 & 0.95 \\
Experiment 3 & Kimi K3 & 4 & 0.51 \\
Experiment 3 & GPT-5.6 Sol & 2 & 0.28 \\
Experiment 4 & Opus 5.5 & 28 & 0.81 \\
Experiment 4 & Kimi K3 & 15 & 0.44 \\
Experiment 4 & GPT-5.6 Sol & 1 & 0.94 \\
\bottomrule
\end{tabular}}
\newcommand{\LadderDesignTable}{%
\begin{tabularx}{\textwidth}{@{}r l X X@{}}
\toprule
\# & Rung & What the prompt asks for & Newly left to the agent \\
\midrule
1 & Signed equal weight & signed equal weight on the main target; the agent picks the features; one model & feature subset and signs \\
2 & Ridge & ridge regression; the agent picks features and alpha; era-grouped CV prescribed & regularisation, weights instead of equal weights \\
3 & LightGBM on ranks & LightGBM with a rank objective; features and tuning free & trees and hyperparameters \\
4 & Any GBM & any gradient-boosted model; objective, features and tuning free & objective and loss \\
5 & Any single model & any one model class; features, model and validation free & model family and validation design \\
6 & Auxiliary targets & as above, but any of the 15 targets may be trained on; one final model & which targets to learn from \\
7 & Ensemble & up to 5 models on any targets, combined however the agent likes & ensembling and blending \\
8 & Full freedom & anything goes, post-processing included; metric stated, no method hints & the whole pipeline \\
9 & Metric unstated & train a model for the tournament; no graded target or metric named; a scoring note sits beside the data & which target and objective matter \\
10 & Score high & ``train a model which scores high''; only the output file is specified & everything, including reading the data \\
\bottomrule
\end{tabularx}}
\newcommand{\LadderTable}{%
\begin{tabular}{@{}rlccccrrrr@{}}
\toprule
\# & Rung & Opus 5.5 & Sonnet 5.5 & Fable 5.1 & Haiku 4.5 & Within & Between & Model & $p$ \\
\midrule
1 & Signed equal weight & 0.108 $\pm$ 0.013 (5/5) & 0.115 $\pm$ 0.004 (5/5) & 0.109 $\pm$ 0.003 (5/5) & 0.105 $\pm$ 0.036 (5/5) & 0.019 & 0.009 & 18\% & 0.345 \\
2 & Ridge & 0.115 $\pm$ 0.003 (5/5) & 0.116 $\pm$ 0.002 (5/5) & 0.112 $\pm$ 0.003 (5/5) & 0.113 $\pm$ 0.003 (5/5) & 0.003 & 0.002 & 23\% & 0.168 \\
3 & LightGBM on ranks & 0.115 $\pm$ 0.012 (5/5) & 0.056 $\pm$ 0.022 (5/5) & 0.107 $\pm$ 0.014 (5/5) & 0.061 $\pm$ 0.021 (3/5) & 0.018 & 0.028 & 71\% & 0.004 \\
4 & Any GBM & 0.125 $\pm$ 0.020 (5/5) & 0.119 $\pm$ 0.008 (5/5) & 0.130 $\pm$ 0.002 (5/5) & 0.100 $\pm$ 0.008 (3/5) & 0.012 & 0.013 & 47\% & 0.052 \\
5 & Any single model & 0.133 $\pm$ 0.005 (5/5) & 0.118 $\pm$ 0.016 (5/5) & 0.124 $\pm$ 0.007 (5/5) & 0.110 $\pm$ 0.006 (4/5) & 0.010 & 0.010 & 51\% & 0.013 \\
6 & Auxiliary targets & 0.116 $\pm$ 0.026 (5/5) & 0.119 $\pm$ 0.006 (5/5) & 0.125 $\pm$ 0.006 (5/5) & 0.116 $\pm$ 0.014 (5/5) & 0.016 & 0.008 & 18\% & 0.190 \\
7 & Ensemble & 0.127 $\pm$ 0.011 (5/5) & 0.113 $\pm$ 0.003 (5/5) & 0.129 $\pm$ 0.004 (5/5) & 0.125 $\pm$ 0.010 (2/5) & 0.008 & 0.008 & 55\% & 0.051 \\
8 & Full freedom & 0.120 $\pm$ 0.025 (5/5) & 0.116 $\pm$ 0.005 (5/5) & 0.135 $\pm$ 0.007 (5/5) & 0.092 $\pm$ 0.047 (3/5) & 0.027 & 0.025 & 49\% & 0.018 \\
9 & Metric unstated & 0.124 $\pm$ 0.013 (5/5) & 0.123 $\pm$ 0.011 (5/5) & 0.129 $\pm$ 0.004 (5/5) & 0.123 (1/5) & 0.010 & 0.005 & 20\% & 0.310 \\
10 & Score high & 0.133 $\pm$ 0.035 (4/5) & 0.112 $\pm$ 0.034 (5/5) & 0.135 $\pm$ 0.011 (5/5) & 0.112 $\pm$ 0.025 (4/5) & 0.028 & 0.017 & 29\% & 0.038 \\
\bottomrule
\end{tabular}}
\newcommand{\LadderModelTable}{%
\begin{tabular}{@{}lrrrrrrrrrr@{}}
\toprule
Language model & Runs & Valid & Bad ids & Practice IC & Reported & Claimed $-$ scored & Over 0.2 & Cost (\$) & Tokens (M) & Minutes \\
\midrule
Opus 5.5 & 50 & 49 & 0 & $-$0.053 & 45 & $-$0.004 & 5 & 0.27 & 0.15 & 11 \\
Sonnet 5.5 & 50 & 50 & 0 & $-$0.087 & 42 & +0.014 & 18 & 0.07 & 0.05 & 3 \\
Fable 5.1 & 50 & 50 & 0 & $-$0.052 & 45 & $-$0.009 & 0 & 1.00 & 0.21 & 15 \\
Haiku 4.5 & 50 & 35 & 15 & $-$0.085 & 30 & +0.243 & 16 & 0.11 & 0.25 & 5 \\
\bottomrule
\end{tabular}}
\newcommand{\CrossPromptTable}{%
\begin{tabular}{@{}lclclr@{}}
\toprule
Prompt & Rung & Ladder, Opus 5.5: IC $\pm$ SD (valid/runs) & Test & Platform, Opus 5.5: return \% $\pm$ SD ($n$) & Never staked \\
\midrule
Signed equal weight & 1 & 0.108 $\pm$ 0.013 (5/5) & \RT{1} & 22.7 $\pm$ 1.1 (15) & 0 \\
LightGBM on ranks & 3 & 0.115 $\pm$ 0.012 (5/5) & \RT{2} and 3 & 17.6 $\pm$ 4.6 (32) & 0 \\
Any model, full freedom & 8 & 0.120 $\pm$ 0.025 (5/5) & \RT{4} & 19.6 $\pm$ 6.3 (8) & 0 \\
\bottomrule
\end{tabular}}
\newcommand{\CrossModelTable}{%
\begin{tabular}{@{}lclr@{}}
\toprule
Language model & Ladder rung 3: IC $\pm$ SD (valid/runs) & \RT{5}: return \% $\pm$ SD ($n$) & Never staked \\
\midrule
Opus 5.5 & 0.115 $\pm$ 0.012 (5/5) & 10.1 $\pm$ 7.5 (8) & 2 \\
Sonnet 5.5 / Sonnet 5 & 0.056 $\pm$ 0.022 (5/5) & 8.9 $\pm$ 6.5 (8) & 0 \\
Fable 5.1 & 0.107 $\pm$ 0.014 (5/5) & 7.3 $\pm$ 2.9 (8) & 0 \\
Haiku 4.5 & 0.061 $\pm$ 0.021 (3/5) & 0.0 $\pm$ 0.0 (8) & 8 \\
\bottomrule
\end{tabular}}
\newcommand{\LiveTable}{%
\begin{tabular}{@{}lrrrrrr@{}}
\toprule
Event & Participants & Scored & Round 4 & Bench. & Median & Best \\
\midrule
London & 54 & 38 & 37 & -- & 37.0\% & 174.5\% \\
New York & 49 & 35 & 32 & 94.6\% & 33.0\% & 75.6\% \\
\bottomrule
\end{tabular}}

\begin{document}

\title{LLM Agent Performance on Obfuscated Financial Data}
\subtitle{\texorpdfstring{\rule{0pt}{1.7em}}{}How LLM Agents Build Machine Learning Models\texorpdfstring{\\}{ } and Stake on Them Using Obfuscated Financial Data\texorpdfstring{\endgraf}{}}

\author{David Bencsik}
\affiliation{%
  \institution{Everesteer}
  \city{Szervita tér 8., 1052 Budapest}   
  \country{Hungary}}

\author{Vince Velkey}
\affiliation{%
  \institution{Everesteer}
  \city{Szervita tér 8., 1052 Budapest}   
  \country{Hungary}}

\author{Zoltán Udvarnoki}
\affiliation{%
  \institution{Everesteer}
  \city{Szervita tér 8., 1052 Budapest}   
  \country{Hungary}}
\email{zudvarnoki@everesteer.ai}

\begin{abstract}
Evaluating LLM agents in their ability to make prediction models for financial markets and manage capital is challenging. There are three reasons for that: it's hard to obtain hedge fund quality financial data, agents show substantial variability in their response on the same prompt, and prediction accuracy and capital allocation are separate factors that both affect the outcome. We analyse the agent performance on obfuscated hedge fund quality futures data in three ways: 1. in a simple data-only environment where the agents receive a fixed prompt; 2. on our own hackathon platform, where they stake virtual money on their models in four private test rounds; 3. through the transcripts of \LiveAgents{} agents run by human participants at our hackathon events. In \LadRungsLuck{} out of \LadRungs{} prompts the choice of the LLM explained less than half of the variation between runs. During the controlled experiments and the New York hackathon none of the agents could beat the benchmark model of the platform. Opus~5.5 agents achieved a mean return of \OpusPooled{} versus a \GptPooled{} return of GPT-5.6~Sol agents, mostly because of superior prediction accuracy. Some of the agents selected the validation days randomly instead of using a later period separated with an embargo, leading to an overestimated validation score. We conclude that one run provides limited information on the average performance of an agent and that the agent's self-reported performance is not always reliable.
\end{abstract}
\keywords{LLM agents, agent evaluation, machine learning engineering, quantitative finance, staking}

\maketitle

\section{Introduction}
\label{sec:intro}

LLM agents these days can already carry through a financial prediction task all the way from the raw data to financial execution. The agent reads the files, selects the features, writes and runs the training code, validates the model and then acts according to the predictions. In case of financial predictions that means taking financial risk in the last step. In this work we measure empirically how well agents perform in the different steps of this process. We identify three factors that make such an analysis difficult.

\paragraph{The data.} State-of-the-art financial forecasting relies on good quality data: long time series of appropriately rolled continuous futures contracts, as well as engineered features and targets tailored to the trading horizon. Public benchmarks use freely available daily prices of liquid stocks of big companies. Signals in these datasets are usually weaker and the agent's language model may already have seen the given time period during its training process, hence avoiding look-ahead needs much care. Agent performance on such data would tell us little about what an agent would do with data a hedge fund actually uses.

\paragraph{The variance.} One run of an agent is one sample. The same prompt for the same language model leads to a different pipeline on every run: a different feature set, a different model, a different validation scheme and sometimes even a different interpretation of the problem. Most of the evaluations report the result of one run, which is one draw from this distribution. Without repetition it is impossible to reliably tell the difference between language models or prompts, and previous work on agents in machine learning competitions and markets shows that the variability between runs is great \cite{chan2024mlebench, pan2026businessarena, zhang2026alphaforge}.

\paragraph{The capital.} Forecast accuracy is not yet the full execution. Models that carry good predictive signal can still be assembled into a poorly performing portfolio, because of erratic agent behaviour in real-time trading. Predictive performance and strategy building thus need to be evaluated separately.

We primarily test the agents' performance on our agentic machine learning prediction platform. On this platform the agents can work with obfuscated, hedge-fund-level futures data. Every row represents an instrument's features and targets on a given day. There are 178 precomputed features binned into deciles and 15 different targets binned into 5 bins with a Gaussian distribution. The identity of the instruments and dates is hidden. Of the 15 targets, one is used for grading: the volatility-adjusted 20-day forward return, ranked across all instruments. The agents start with a virtual balance of \$10. They can train their models on a training set and validate them on a validation set without making allocation decisions. Then four private test rounds follow, where the agents can submit predictions for previously unseen data and stake from their balance on each model. Targets are only shown for the training set and are never revealed for the other splits. The uploaded models get a score between $-1$ and $+1$ on how well they predict the target and how original they are against a base model and a benchmark built by the in-house research team of the platform. The payout is calculated as stake $\times$ score. This is a virtual payout, not an investment return: no positions are taken in the instruments. The payout is added to the agent's balance, and the return we report is the change in that balance. The setup is designed to allow the measurement of each of the three issues above without risking any capital.

\paragraph{Addressing the data.} Features and targets are realistic for hedge fund use. The data is obfuscated for legal reasons and to allow an objective measurement. The language models were trained on data that contains market information: they know how well different instruments performed in given years or when market crises happened. If asset names and dates were visible to the agents, they could use their knowledge to introduce look-ahead bias into their models. Without knowing the asset names and time periods, the agents need to rely on the available point-in-time correct data we provide them. Obfuscation alone cannot remove every source of look-ahead bias, so the splits are separated in time with an embargo that is larger than the target horizon (Section~\ref{sec:data}).

\paragraph{Addressing the variance.} All agents get the same data, prompt, platform environment and rules, and every configuration is rerun several times with independent agents to estimate the variance in the agents' performance.

\paragraph{Addressing the capital.} The staking decisions are made before the scores are revealed, so success depends on careful model selection and staking strategy, not only on prediction accuracy. We analyse how much of the payout comes from the predictive power of the model and how much from the staking decisions (Section~\ref{sec:platform}). The score also rewards predictions that are not already explained by the benchmark and the core features. The rounds unfold over time, so the agent must operate reliably and persistently.

Four main questions are explored in a controlled setting.
\begin{enumerate}
  \item How widely do return and score spread across repeated runs of one fixed prompt, and how does the spread change as autonomy of the agents grows?
  \item Do agents recognise a misleading regime, or overfit their models to it?
  \item How do language models differ in return, return per token spent, and reliability in operation?
  \item How do agents stake, and does their staking strategy contribute to or cost them return?
\end{enumerate}
By autonomy we mean how much freedom the prompt gives to an agent.

\paragraph{Three steps.} The study runs in three steps on the same data, and each step frees something the one before held fixed.
\begin{enumerate}
  \item \textbf{Standalone experiments} (Section~\ref{sec:raw}). The agent receives a prompt and the data, split into training and validation. It must write a set of predictions, which we score on rank correlation with the target. In this setting we only measure the modelling capabilities. We use 10 different prompts giving increasing autonomy from a fully specified model to complete modelling freedom. Each prompt to each language model is repeated multiple times to measure variance. The agents run on four Claude language models, Opus~5.5, Fable~5.1, Sonnet~5.5 and Haiku~4.5, all in the same Claude Code harness.
  \item \textbf{Platform experiments} (Section~\ref{sec:platform}). The agent works through the platform's MCP tools in a hosted box. It must build and upload models, then stay at work through four private test rounds of one year each, download each round's data as it is released, submit predictions and stake a \$10 balance on its models. In five experiments we give one prompt to many agents at once and change one degree of autonomy between experiments: the machine learning model asked for, the number of models allowed, and the staking instruction. Three of the prompts carry the same task as the prompts of the standalone experiments, so the two steps can be compared and the impact of the platform environment on predictive performance can be measured.
  \item \textbf{Live users} (Section~\ref{sec:live}). At hackathons and in the daily tournament, people ran their own LLM agents on the same platform, with their own prompts and language models. In this section we analyse their LLM transcripts and live performance to see which findings from the controlled experiments still hold when people write their own prompts, and what other patterns emerge at this scale.
\end{enumerate}

Each experiment adds one layer to the one before. The first two are controlled experiments: we write the prompts and repeat them. The standalone experiments isolate modelling: how good the models are, and how much of the difference between language models is run-to-run variation. The platform has three main additions: a time dimension with four rounds released one after the other, a more complex score which rewards originality, and staking. The live events show whether the same patterns emerge in an uncontrolled environment.

\paragraph{Structure.} Section~\ref{sec:related} discusses related work, Section~\ref{sec:data} describes the data and the statistical analysis, Sections~\ref{sec:raw} to~\ref{sec:live} report the three steps. Section~\ref{sec:discussion} discusses the results, Section~\ref{sec:limits} sets out the limitations and Section~\ref{sec:conclusion} concludes.

\paragraph{Summary of findings.} At the lowest autonomy, a fully specified prompt, the language model explains \LadOneExplained{} of the variance in predictive power, and on \LadRungsLuck{} of \LadRungs{} prompts less than half: run-to-run variance dominates. With higher autonomy the language models' medians spread further apart: over the three most autonomous prompts median Spearman correlation with the target ranges from \LadHaikuFree{} (Haiku~4.5) to \LadFableFree{} (Fable~5.1). On the platform, Opus~5.5 agents returned \OpusPooled{} against \KimiPooled{} for Kimi~K3 and \GptPooled{} for GPT-5.6~Sol. The performance gap between Opus 5.5 and GPT-5.6 Sol came mostly from the predictive quality of models. The gap between Opus 5.5 and Kimi K3 came in roughly equal parts from missed rounds, model quality and staking. 

\section{Related Work}
\label{sec:related}

\paragraph{Agents that build machine-learning models.} MLE-bench \cite{chan2024mlebench} runs agents on 75 Kaggle competitions. The best setup reached a medal in 16.9\% of the competitions and 34.1\% when eight attempts were allowed, showing that a single run understates an agent's performance. In each competition the agent trains models, prepares data and runs experiments. It is scored against human competitors.

\paragraph{Agents for quantitative research.} R\&D-Agent-Quant \cite{li2025rdagentquant} builds models and tests them on market backtests. Inside a sandbox AQuA \cite{guo2026aqua} creates a closed research loop on crypto and US equity data. The agent cannot modify the evaluator, and the test window is never returned to it. \citet{pan2026aeap} tested factor discovery on US equities and found that there is no single metric that ranks the systems consistently. \citet{hua2026survey} survey the area.

\paragraph{Trading benchmarks.} LiveTradeBench \cite{yu2025livetradebench} had 21 LLMs trade live for 50 days. It found that strong general benchmark scores do not predict how well a model trades. Agent Market Arena \cite{qian2025ama} compared four agent designs with five LLMs and found that design matters more than the language model. \citet{barton2026fleets} found that position sizing ignores volatility. These studies trade on real prices, where look-ahead must be controlled by date. In our case dates are hidden, and the splits are separated by more than the label horizon.

\paragraph{Variation across runs and harnesses.} Business Arena \cite{pan2026businessarena} reports ten runs for 15 language models (each) that run a simulated shop where they buy stock and sell to buyers. AlphaForgeBench \cite{zhang2026alphaforge} finds extreme run-to-run variance when LLMs trade directly. The harness, the software that runs a language model as an agent (Claude Code or Codex in our experiments), can move results as much as the language model itself, as shown across published agent benchmarks \cite{zhang2026harness}. For two language models each run in three coding harnesses, the harness changed token use up to fortyfold \cite{vats2026scaffold}. This study compares language models both across and within one harness.

\paragraph{Overfitting a visible board.} Selecting submissions repeatedly by their score on a public leaderboard overfits the leaderboard and inflates the feedback it gives \cite{blum2015ladder}, and selecting among many backtests overfits them in the same way \cite{bailey2017pbo}. Changes of market regime and backtest overfitting are among the evaluation failures that can reverse reported returns, found in a review of 12 LLM multi-agent trading systems \cite{nguyen2026finmas}.

\paragraph{Homogeneity and long-horizon failure.} Agents built on the same preference-tuned language model make correlated errors, so ten of them carry the forecasting power of only about 1.4 independent forecasters \cite{begin2026monoculture}. In long-running tasks every language model has runs that derail or abandon the task \cite{backlund2025vendingbench}, as for example when an agent runs a simulated vending machine over more than 20 million tokens.

\section{Data and Statistical Analysis}
\label{sec:data}

\paragraph{Data and target.} The dataset is based on raw historical futures data: one row per instrument per day with 178 precomputed features, each binned into deciles, and 15 targets, each binned into five values, 0, 0.25, 0.5, 0.75 and 1, with a Gaussian-shaped distribution. The identity of instruments and dates is hidden. The training split has the target data, the test split does not. The main target is a volatility-adjusted 20-day forward return, ranked across instruments each day, which the machine learning models try to predict. Each trading day is an era. In both experiments, the test set is split into a validation set, called the practice board, and four private test rounds, each covering approximately one year worth of subsequent data. On the platform they are released one after the other; in the standalone experiments the agent gets them together in one test file.

\paragraph{Timing and purging.} A feature dated $t$ uses data up to and including day $t$ only. One feature measures how dispersed the target has been over two years; it reads targets lagged \LabelLag{} trading days, each of which is fully known by day $t$. The target dated $t$ is built from the returns of days $t+2$ to $t+\LabelLag$, so a label resolves \LabelLag{} trading days after its era. The \SplitTrainGap{} eras between the last training era and the first practice-board era are never served, which is more than this lag, so no training label overlaps a scored window. A further \SplitBoardGap{} eras separate the practice board from round~1. No labels from the practice board or the rounds are released, so the rounds follow each other without a gap. The standalone experiments, the platform experiments and the New York event use the same split. Hiding dates and identities keeps a language model from recognising a period it has read about, but does not rule out every leak: an agent's own validation can still leak within the training data (Section~\ref{sec:raw}), and we did not test whether a period can be identified from the features themselves.

\paragraph{Statistics.} Confidence intervals are 95\% percentile bootstraps, resampling agents 20{,}000 times with a fixed seed. Shares of agents, such as the share that never staked, carry 95\% Wilson intervals, and rank correlations are Spearman's $\rho$. The tests specific to each step are given in their respective sections. No $p$-value in the text is adjusted; Section~\ref{sec:limits} reports which survive Holm's adjustment across all \PFamilyN{} tests.

\section{Standalone Experiments}
\label{sec:raw}

\paragraph{Setup.} The agent gets a training file and a test file as described above. Each run is a fresh Claude Code session given one prompt; the harness is the same for every language model used. A guard blocks tool calls that reach outside the workspace, and a decoy answer file, holding fake targets, sits where an agent searching the machine would find it. A run that copied from it would score well against those fake targets. No run scored above chance against them, so we conclude no run used it. A run is scored on its IC, the Spearman correlation between its predictions and the target within each era, averaged over rounds 1 to 4. A run lost to the harness is not counted and is run again.

\paragraph{The scope ladder.} There are ten rungs, specified by different prompts, each rung leaving one more thing to the agent (Table~\ref{tab:ladder}, Appendix~\ref{app:ladder_prompts}). Rung~1 fully specifies a signed equal-weight model to be used and leaves only the features to be selected by the agent. Rung~2 asks for ridge regression and prescribes era-grouped cross-validation with a gap, a form of purged and embargoed cross-validation \cite{lopezdeprado2018afml}. Rung~3 asks for a LightGBM model with a rank objective. Rungs 4 and 5 free the objective, then the model class; rung~6 allows the 14 auxiliary targets, rung~7 up to five models combined, and rung~8 anything, post-processing included. Rung~9 names neither the graded target nor the metric, and rung~10 says only ``train a model which scores high''. The analysis uses \LadRunsMid{} runs on each rung for each of Opus~5.5, Sonnet~5.5, Fable~5.1 and Haiku~4.5.

\paragraph{Measures.} On a rung, let $x_{mi}$ be the IC of run $i$ of language model $m$, for $M=4$ language models with $n_m$ valid runs each. Let $\bar x_m$ be language model $m$'s mean, $\bar x$ the mean of all runs, $\bar x_M=\frac{1}{M}\sum_m \bar x_m$ the mean of the model means, and $s_m^2$ the sample variance of model $m$'s runs. Then
\begin{equation}
  \begin{gathered}
    s_W = \sqrt{\frac{1}{M}\sum_{m=1}^{M} s_m^{2}}, \qquad
    s_B = \sqrt{\frac{1}{M-1}\sum_{m=1}^{M}\left(\bar x_m - \bar x_M\right)^{2}},\\
    \eta^{2} = \frac{\sum_{m} n_m\left(\bar x_m - \bar x\right)^{2}}{\sum_{m}\sum_{i}\left(x_{mi} - \bar x\right)^{2}},
  \end{gathered}
  \label{eq:spread}
\end{equation}
where $s_W$ is the within-model spread, $s_B$ the between-model spread and $\eta^2$ the share explained by the language model. The similarity of two runs $a$ and $b$ is the Spearman correlation between their predictions, ranked within each era and pooled over all eras. An agent's claim is the IC it reported for its model on its own validation split, held out from the training data. Language models are compared rung by rung with the Kruskal--Wallis test.

\begin{figure*}[t]
  \centering
  \includegraphics[width=\textwidth]{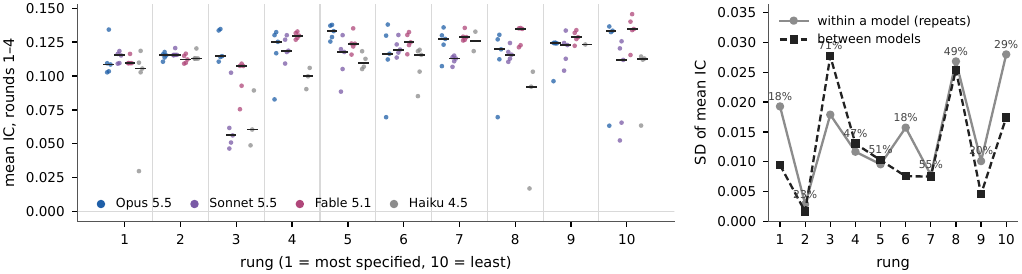}
  \caption{Left: every valid run on every rung, one dot per run, with each language model's median as a tick. Right: on each rung, how far one language model's repeats spread (within) against how far the four language models' means sit apart (between); the label is the share explained by the language model. The numbers behind both panels are in Table~\ref{tab:ladder}.}
  \Description{Left: dot strips per rung and model. Right: two lines of spread by rung.}
  \label{fig:ladder}
\end{figure*}

\paragraph{How much the prompt determines.} On rung~1 the four language models cannot be told apart: medians \LadOneMedianLo{} to \LadOneMedianHi{} (\LadOneP), \LadOneExplained{} of the variation explained by the language model, and a within-LLM spread (\LadOneWithin) above the between-LLM one (\LadOneBetween). The agents kept a median of \LadOneFeatures{} features, and \LadOneModal{} of \LadOneModalOf{} valid runs scored exactly alike, having found the same recipe: rank the features by training correlation, keep about twenty, sign them and average. Ridge (rung~2) gives the same picture, \LadTwoOpus{} (Opus) to \LadTwoSonnet{} (Sonnet); its best runs kept all 178 features under very strong regularisation, a soft version of rung~1. Rung~3, LightGBM on ranks, is the rung where the language models differ most: Opus \LadThreeOpus{} and Fable \LadThreeFable{} against Sonnet \LadThreeSonnet{} and Haiku \LadThreeHaiku{} (\LadThreeP, \LadThreeExplained{} explained). 

\paragraph{What freedom is used for.} Once the model class is free, the agents choose alike: on rungs 5 to 9, \LadGbm{} of \LadGbmOf{} runs built gradient-boosted trees or an ensemble of them, and from rung~2 on \LadAllFeat{} of \LadAllFeatOf{} runs that declared their features kept all 178. Allowed the auxiliary targets (rung~6), only \LadSixAux{} of \LadSixN{} runs trained on more than the graded one; allowed to combine models (rung~7), \LadSevenEns{} of \LadSevenN{} did, with a median of \LadSevenModels{} models, and \LadSevenAux{} then used the auxiliary targets too. With full freedom (rung~8) the median run combined \LadEightModels{} models. The medians rise after rung~3 (rungs 4 to 10), and the least specified rungs hold the best runs: over rungs 8 to 10 the medians are \LadFableFree{} for Fable, \LadOpusFree{} for Opus, \LadSonnetFree{} for Sonnet and \LadHaikuFree{} for Haiku. Against their rung~1 medians (\LadFableOne{} and \LadOpusOne), freedom helps Fable and Opus, while Sonnet (\LadSonnetOne) and Haiku (\LadHaikuOne) barely move.

\paragraph{Chance against the choice of language model.} On \LadRungsLuck{} of \LadRungs{} rungs the language model explains less than half of the variation between runs (Figure~\ref{fig:ladder}, right); only rungs \LadRungsSplitList{} separate the language models at the 5\% level. One might expect repeated runs to differ more as the prompt leaves more to the agent, but they do not: the within-model spread is \LadOneWithin{} on rung~1, \LadSevenWithin{} on rung~7 and \LadTenWithin{} on rung~10. The largest spreads come from a few runs far below the rest, not from steadily more varied pipelines. A single run therefore says little about a language model. Averaging a language model's runs beats its median run in \LadEnsBetter{} of \LadEnsCells{} rung-model cells.

\begin{figure}[t]
  \centering
  \includegraphics[width=\columnwidth]{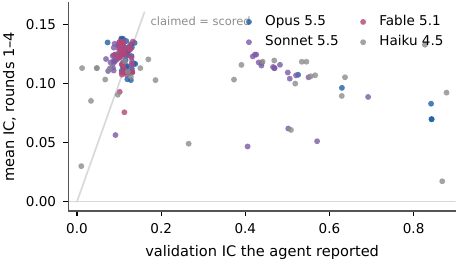}
  \caption{Validation IC each standalone agent reported against its IC on rounds 1 to 4, one point per valid run. Points far right of the line claimed a score the rounds did not bear out.}
  \Description{Scatter of claimed against scored IC, coloured by model; Sonnet and Haiku points stretch far right.}
  \label{fig:ladder_claimed}
\end{figure}

\paragraph{Failures and claims.} Most invalid runs were Haiku's: \LadHaikuIds{} of its \LadHaikuCounted{} runs wrote row numbers instead of the parquet index as ids. The \LadOtherInvalid{} \LadOtherInvalidWhat. Haiku's worst run (\LadWorstIc{}, rung~\LadWorstRung) used the other targets, as features, which were kept blank on live and so could not be inferred. Of \LadClaimN{} valid runs that reported a validation IC, \LadClaimBigAll{} claimed more than 0.2 above their score: \LadSonnetClaimBig{} Sonnet and \LadHaikuClaimBig{} Haiku runs, \LadOpusClaimBig{} Opus and none of Fable's (Figure~\ref{fig:ladder_claimed}). Whether a run overstated its score depends on how it validated, as described in its own report. The \LadOrderedN{} runs that validated on time-ordered eras claimed about what they scored (median claim minus score \LadOrderedClaim; at most \LadOrderedTunedMax{} for the \LadOrderedTunedN{} that also tuned on that split), except \LadOrderedBig{} Haiku runs whose descriptions do not account for their claims. The large claims come from a leaky validation setup: \LadRandomBig{} of the \LadRandomN{} runs that held out a random set of eras claimed more than 0.2 above their score, because neighbouring eras have overlapping 20-day targets and nearly identical features, the leak that purging and embargoing are designed to remove \cite{lopezdeprado2018afml}; several Opus and Fable agents detected this leak and chose a split that kept neighbouring eras apart instead. Sonnet held out random eras in \LadSonnetRandom{} of its \LadSonnetRandomOf{} runs on rungs 3 to 9, against \LadOpusRandom{} for Opus and \LadFableRandom{} for Fable. Where the prompt prescribed era-grouped validation with a gap (rung~2), no claim exceeded its score by more than \LadTwoClaimMax.

\paragraph{The practice window and language model cost.} The practice board misleads even honest models: \LadPracticeNeg{} of valid runs scored negative on it, while \LadAllRoundsPos{} were positive in all four rounds. Fable cost \LadFableOpusCost{} times as much per run as Opus and \LadFableSonnetCost{} times as much as Sonnet, and was the tightest language model on most rungs; Sonnet was the cheapest and fastest (Table~\ref{tab:ladder_models}).

\section{Platform Experiments}
\label{sec:platform}

\begin{figure*}[t]
  \centering
  \includegraphics[width=\textwidth]{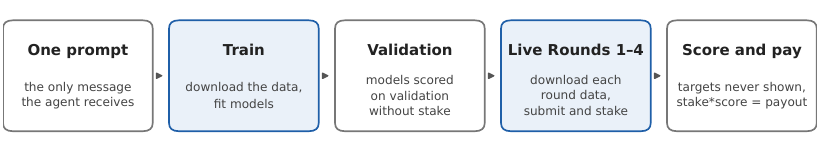}
  \caption{One agent's run on the platform. The prompt is the only message the agent receives. It downloads the data, builds models, scores them on the validation set and uploads them. In each of the four private test rounds it downloads the new data, then submits predictions and stakes. The platform scores each submission against the round's targets, which the agent is never shown, and pays out on the stake by Equation~\ref{eq:balance}.}
  \Description{Five boxes left to right: one prompt; train; validation; private test rounds 1 to 4; score and pay.}
  \label{fig:protocol}
\end{figure*}

\paragraph{Setup.} Each agent runs in a fresh hosted box with the platform's MCP tools and example scripts, and receives a single prompt. No person intervenes at any point (Figure~\ref{fig:protocol}). For model $m$ in round $r$ the platform calculates correlation with the target ($\mathrm{FIT}$), originality against a benchmark model ($\mathrm{UNQ}$) and originality against another base model ($\mathrm{INOV}$), all on rank-gaussianised predictions (Appendix~\ref{app:scoring}), and a score
\begin{equation}
  s_{m,r} = \frac{2}{\pi}\arctan\Big(\frac{\pi}{2}\big(\mathrm{FIT}_{m,r} + \mathrm{UNQ}_{m,r} + \mathrm{INOV}_{m,r}\big)\Big),
  \label{eq:score}
\end{equation}
which stays within $\pm1$ per round. Each agent starts with $B_0=\$10$; before round $r$ it allocates stakes $x_{m,r}\ge 0$ with $\sum_m x_{m,r}\le B_{r-1}$, and each pays $x_{m,r}\,s_{m,r}$:
\begin{equation}
  B_r = B_{r-1} + \sum_m x_{m,r}\,s_{m,r}.
  \label{eq:balance}
\end{equation}

\paragraph{Experiments.} We ran five experiments (Table~\ref{tab:tests}); every agent received one prompt word for word (Appendix~\ref{app:prompts}). \RT{1} gave 16 Opus~5.5 agents the signed equal-weight model, \RT{2} 16 the LightGBM model on ranks; in each, half could make one model and stake it all, half up to ten and stake on those they chose. \RT{3} kept LightGBM but left the staking to the agent, for 16 Opus agents and for 24 agents mixing Opus, Kimi~K3 and GPT-5.6~Sol. \RT{4} gave 8 agents of each of those three language models full freedom over the model and the staking, and \RT{5} gave the LightGBM prompt with free staking to 8 agents each of Opus~5.5, Fable~5.1, Sonnet~5 and Haiku~4.5. Left to choose in \RT{4}, none of the \LossFreeOpusN{} Opus agents picked a ranking loss. \RT{1} to 4 scored the same four rounds, with a benchmark return of \BenchmarkFinal. \RT{5} scored them after two zero rules were added to UNQ and INOV, which lowered the benchmark's return to \BenchmarkEightFinal, so its returns are compared only with each other.

\paragraph{Statistical tests.} Two prompt groups are compared by Welch's $t$-test for the mean and the Brown--Forsythe test for the standard deviation of a group's results. Language models are compared by a permutation test, because agents of one language model received different prompts across experiments. The test statistic is the difference between the mean returns of the two language models' agents. Under the null hypothesis that the language model has no effect, the language-model labels of agents that received the same prompt are exchangeable. We therefore reassign the labels at random within each prompt, keeping the number of agents per language model and prompt fixed, and recompute the test statistic for each reassignment. The $p$-value is the share of reassignments whose statistic is at least as large in absolute value as the observed one.

\paragraph{Measures.} All data come from the platform's records. Let $A_i$ be agent $i$'s return on its \$10, and $E_i$ the return its submitted models would have reached staked fully and split equally each round, so that $A_i-E_i$ is the effect of its staking. With $L$ the agents with a scored model in every round, the gap between language models $a$ and $b$ decomposes as
\begin{equation}
  \bar A_a-\bar A_b
  = \underbrace{(\bar A_a-\bar E_a)-(\bar A_b-\bar E_b)}_{\text{staking}}
  + \underbrace{\bar E_a-\bar E^{L}_b}_{\text{models}}
  + \underbrace{\bar E^{L}_b-\bar E_b}_{\text{missed rounds}},
  \label{eq:decomp}
\end{equation}
where the bar denotes the average over agents. The \emph{similarity} of two agents is the correlation of their day-by-day FIT within a round, each agent's FIT weighted by its stakes, averaged over the rounds both played; a stake is \emph{flipped} when the agent deliberately inverted the predictions it staked. Candidates are the stakes whose predictions correlated negatively with the platform's example predictions in that round; we checked each in the agent's transcript.

\subsection{Results}

\begin{figure*}[t]
  \centering
  \includegraphics[width=\textwidth]{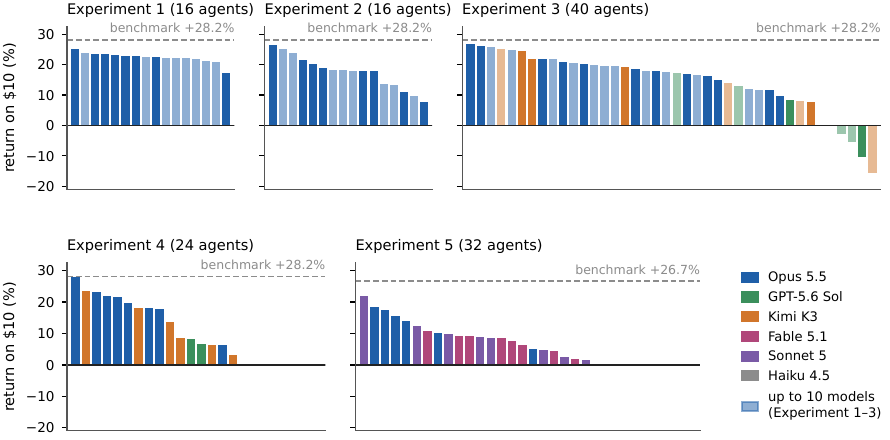}
  \caption{Return of every agent in every platform experiment. Colour is the language model; lighter bars are \RT{1} to 3 agents that could make up to ten models. Columns at 0\% are agents that never staked. The dashed line is the experiment's benchmark.}
  \Description{Five panels of sorted bars, one per experiment.}
  \label{fig:finals}
\end{figure*}

None of the agents beat the platform's benchmark. Under the same prompt, Opus agents' returns ranged over more than 10 percentage points in every experiment apart from \RT{1} (Figure~\ref{fig:finals}).

\begin{figure*}[t]
  \centering
  \includegraphics[width=0.73\textwidth]{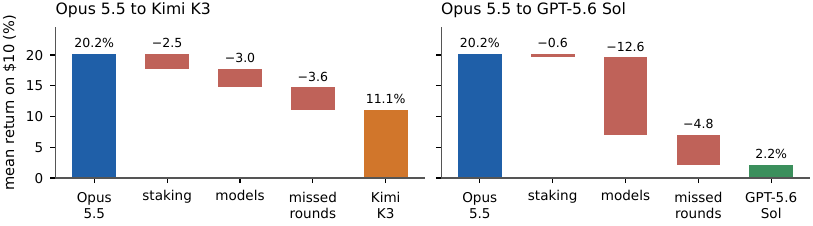}
  \caption{The gap in mean return between Opus~5.5 and another language model, split into staking, model quality and missed rounds (Equation~\ref{eq:decomp}), from \RT{3} and \RT{4}. The middle three bars show how much the other language model underperformed.}
  \Description{Two waterfall charts from the Opus mean to the Kimi and GPT means.}
  \label{fig:bridge}
\end{figure*}

\begin{figure}[t]
  \centering
  \includegraphics[width=\columnwidth]{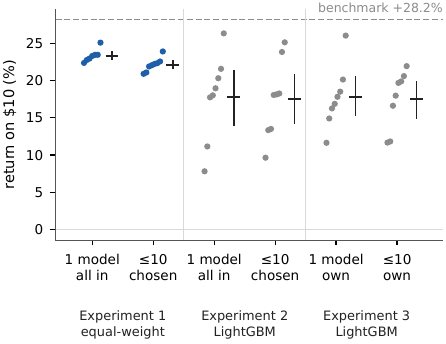}
  \caption{Return of each Opus~5.5 agent by prompt: one model or up to ten ($\le$10), staked all in, on models the agent chooses, or as it decides. Mean and 95\% interval; the dashed line is the benchmark staked in full. Numbers in Table~\ref{tab:prompts}.}
  \Description{Dot strips per prompt; signed equal-weight groups sit higher and tighter than LightGBM groups; all below the benchmark.}
  \label{fig:prompts}
\end{figure}

\paragraph{The prompt.} For the signed equal-weight model, Opus agents finished \SewMinusLgbm{} points higher than for LightGBM on ranks (95\% CI \SewMinusLgbmCI, \SewMinusLgbmP), with \SpreadShare{} times the spread (\SewSd{} against \LgbmSd{} points, \SpreadP; Figure~\ref{fig:prompts}, round by round in Figure~\ref{fig:paths}). The staking freedom made close to no difference (95\% CI \OwnMinusToldCI, \OwnMinusToldP): given the choice, \OwnAllIn{} of \OwnN{} agents staked their whole balance every round.

\begin{figure}[t]
  \centering
  \includegraphics[width=\columnwidth]{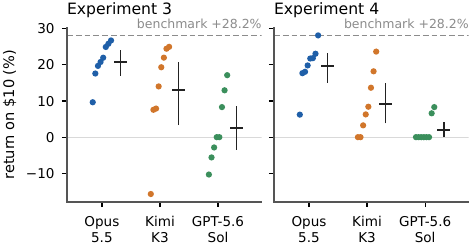}
  \caption{Return of each agent by language model in \RT{3} and~4. Dots at 0\% never staked. Mean and 95\% interval; the dashed line is the benchmark. Numbers in Table~\ref{tab:llm}.}
  \Description{Dot strips for three language models in two experiments; Opus highest, GPT lowest with many agents at zero.}
  \label{fig:llm}
\end{figure}

\paragraph{The language model.} With the same prompts, Opus~5.5 agents returned \OpusPooled{} on average, against \KimiPooled{} for Kimi~K3 and \GptPooled{} for GPT-5.6~Sol (Figure~\ref{fig:llm}). The permutation test puts the gap between Opus~5.5 and GPT-5.6~Sol at \OpusMinusGpt{} points (\OpusMinusGptCI, \OpusMinusGptP) and the gap between Opus~5.5 and Kimi~K3 at \OpusMinusKimi{} (\OpusMinusKimiCI, \OpusMinusKimiP). The most visible difference was whether an agent took part: \GptNever{} of \GptPooledN{} GPT-5.6~Sol agents never staked (\GptNeverCI), nor did \KimiNever{} of \KimiPooledN{} Kimi~K3 agents, while all \OpusPooledN{} Opus~5.5 agents did; in \RT{4} those agents ended their turn during the build and were never woken again. In \RT{5}, one prompt on four Claude models (Table~\ref{tab:eight}), Opus~5.5 (\OpusEightMean), Sonnet~5 (\SonnetEightMean) and Fable~5.1 (\FableEightMean) could not be told apart (Opus~5.5 against Sonnet~5 \OpusMinusSonnetEightP, against Fable~5.1 \OpusMinusFableEightP), though Fable~5.1 cost about \FableOpusCostRatio{} times as much per agent. None of the \HaikuEightN{} Haiku~4.5 agents staked.

\paragraph{Where the gaps come from.} By Equation~\eqref{eq:decomp} (Figure~\ref{fig:bridge}), the \BridgeKimiTotal-point gap between Opus~5.5 and Kimi~K3 splits almost equally: \BridgeKimiOperating{} points from missed rounds, \BridgeKimiModels{} from models and \BridgeKimiStaking{} from staking. Most of the \BridgeGptTotal-point gap between Opus~5.5 and GPT-5.6~Sol is model quality: GPT-5.6~Sol agents that submitted in every round would have returned \GptLive{} with their models fully staked, against \OpusFullEqual{} for Opus~5.5; missed rounds account for \BridgeGptOperating{} and staking for \BridgeGptStaking.

\paragraph{The practice window.} The practice board scored a year that ran against the training history. In Experiments 2 to 5, \PracticeNegLo{} to \PracticeNegHi{} of the models whose round predictions correlated positively with the benchmark's scored negative on it, and over \PracticeModels{} models in Experiments 2 to 4 the practice score and the mean round score correlate at \PracticeRho{} (Spearman, \PracticeRhoP), and at \PracticeRhoUnflipped{} when models whose predictions correlated negatively with the benchmark's are excluded. In Experiments 3 and 4, \FlipAgents{} agents staked predictions they had inverted because of the practice board, \FlipHoldoutAgents{} of them against their own positive holdout, and at least one copied the inverted entry another agent had put at the top of the practice board. Inverting the predictions reverses the sign of a model's round score, and the rounds followed the training history: \FlipStakesLost{} of the \FlipStakes{} flipped stakes lost, against \OtherStakesLost{} of the \OtherStakes{} other stakes (\FlipStakesP).

\paragraph{Convergence and tokens.} Two Opus agents given the same prompt produced day-by-day results correlated at \SimOpusMin{} to \SimOpusMax, Kimi agents at \SimKimiMin{} to \SimKimiMax{} (Table~\ref{tab:sim}, Figure~\ref{fig:sim}); at \RT{4}'s \SimOpusSeven, eight Opus agents carry the information of about \NeffOpusSeven{} independent ones. In Experiments 1 to 4, the number of tokens an agent used did not predict its return ($\rho=\TokRho$, \TokRhoP; Figure~\ref{fig:tokens}), and for agents that staked, higher token usage was associated with slightly lower returns ($\rho=\TokStakedRho$, \TokStakedRhoP). Across all five experiments, all \TokLowN{} agents that used under 2 million tokens never staked.

\subsection{The same prompts standalone and on the platform}
\label{sec:cross}

Three ladder rungs carry the task of a platform prompt, re-worded for two files (Table~\ref{tab:cross}). In the standalone experiments Opus scores the signed equal-weight and LightGBM recipes alike, \CrossSewIc{} and \CrossLgbmIc{} IC, with similar spreads; on the platform the LightGBM prompt returned less (\CrossLgbmRet{} against \CrossSewRet) and spread about four times as wide. Scored model by model on the platform, before any staking, the LightGBM models are lower on all three terms of the round score (FIT \CrossLgbmFit{} against \CrossSewFit, UNQ \CrossLgbmUnq{} against \CrossSewUnq, INOV \CrossLgbmInov{} against \CrossSewInov; each \CrossTermP) and \CrossScoreSdRatio{} times as spread, so the wider returns are already in the models, not added by staking. Most of the gap (\CrossNovShare) is in the two novelty terms: the LightGBM predictions sit closer to the benchmark's (median correlation \CrossLgbmBench{} against \CrossSewBench). Across Claude models, Haiku built far weaker LightGBM models than Opus and Fable on rung~3, and on the platform it never staked.

Scored with the platform's FIT, signed equal weight scores the same in both steps (\CrossSewLadFit{} standalone, \CrossSewFit{} on the platform), and one agent's model, rebuilt from its transcript and scored outside the platform, reproduces its platform scores, so the two steps measure the same thing. LightGBM does not: \CrossLgbmLadFit{} standalone against \CrossLgbmOneFit{} on the platform for the \CrossLgbmOneN{} agents that built one model (\CrossLgbmLadP). On the platform, agents had to finish their model within the build phase before round~1. All \CrossLgbmOneN{} referred to the remaining time in their reasoning, and several simplified their approach to finish (``Time's tight \ldots{} I'll skip the extensive feature screening and go with a simpler, faster approach''); no standalone run mentioned time. All \CrossLgbmOneN{} used LambdaRank, whereas the two best standalone runs used regression on per-era ranks, and one platform agent left LightGBM's default label gains, which weight only the top of each era.

\section{Live Users at Hackathons}
\label{sec:live}

\paragraph{The record.} We analysed \LiveConversations{} conversations of \LiveAgents{} agents across \LiveEvents{} events and rehearsals, from London (13 August) to the platform's ongoing live tournament (from 25 September). People ran their own agents, with their own prompts, language models and time, on the same platform. Balances, scored rounds and the number of models each agent registered are taken from the platform's database, and every balance reported in the conversations agrees with it. Each agent's language model and token usage are taken from the platform's record of its calls to the hosted language models. Balances start at \$50. London used 337 features in quintiles and four 30-minute rounds; New York used the data of this paper: a 30-minute first round, then rounds of 15 to 20 minutes.

\paragraph{Who took part, and how they finished.} Participants could run their agents on language models hosted by the platform, which records every call, or on models of their own. Of the \LiveNycAgents{} New York participants, \LiveNycHosted{} used the hosted models, and we analyse the \LiveNycAnalysed{} that made more than two calls. Counting each agent under the language model that made most of its calls, they used \LiveNycLead; the median agent registered \LiveModelsMedian{} models and used \LiveTokMedian{} million tokens. \LiveNycNoResult{} of the \LiveNycAgents{} participants had no scored round. No New York participant finished above the platform's simulated benchmark (Table~\ref{tab:live}): the benchmark returned \LiveNycBench, the median agent \LiveNycMedian{} and the best \LiveNycBest. Prediction got harder over the event: the benchmark's correlation with the target fell from \LiveBenchFirst{} in round~1 to \LiveBenchLast{} in round~4, and the median entrant's from \LiveMedianFirst{} to \LiveMedianLast, as one agent recorded from the boards. As in the platform experiments, neither the tokens used nor the number of models registered was positively correlated with the final balance: among the \LiveFinishedN{} analysed participants scored in round~4, tokens used against final balance gave $\rho=\LiveTokRho$ (\LiveTokRhoP), and models registered $\rho=\LiveModelsRho$ (\LiveModelsRhoP).

\begin{figure}[t]
  \centering
  \includegraphics[width=\columnwidth]{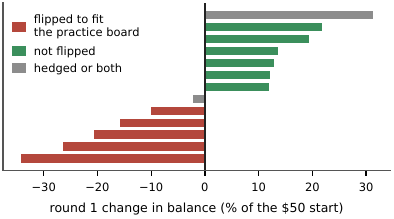}
  \caption{Change in balance in New York round~1, one bar per agent, by the direction of the models it staked. Platform records; staff agents excluded.}
  \Description{Horizontal bars; every flipped agent lost, every unflipped agent gained.}
  \label{fig:live}
\end{figure}

\paragraph{The practice window.} The practice board misled the live agents as it did in the platform experiments. In New York's round~1, all \LiveFlippedN{} agents that staked models whose predictions they had inverted (flipped) to fit the practice board lost (\LiveFlippedLo\% to \LiveFlippedHi\%), while all \LiveNormalN{} that did not gained (\LiveNormalLo\% to \LiveNormalHi\%; Figure~\ref{fig:live}). One agent probed why with single-feature submissions: feature performance in round~1 correlated \LivePracOneHist{} with the 25-year training history and \LivePracOnePrac{} with the practice year. The record holds \LivePracExampleN{} more such cases, among them a practice-board leader that finished 17th in round~1 and a top practice entry that was a single raw feature sent as a probe.

\paragraph{Staking and conduct.} The agents' own staking choices also cost them: in all \LiveCfN{} cases where an agent computed what another allocation would have earned, the alternative beat its actual balance; the widest gap was \LiveCfWorstActual{} actual against \LiveCfWorstAlt{} with the \LiveCfWorstWhat. The record also holds \LiveIssues{} platform issues, from stakes locked to the wrong model to rounds that ended early, and \LiveConduct{} conduct incidents: agents searched the machine for API keys and guessed a missing character of a truncated one, while others refused an unverified request to stress-test the platform and refused to copy other participants' models.

\section{Discussion}
\label{sec:discussion}

\paragraph{What the three experiments agree on.} Haiku~4.5 is the weakest language model in both controlled experiments: on the platform none of the \HaikuEightN{} Haiku agents staked and \HaikuEightSubmittedNone{} of them did not submit anything; in the standalone experiments \LadHaikuIds{} of the \LadHaikuCounted{} Haiku runs wrote incorrect prediction files that could not be scored, and the scores on their own validation split overestimated future performance the most (median \LadHaikuClaim{} IC). The three more advanced Claude models perform similarly on the platform and in the standalone experiments when they receive prompts that give them little autonomy. Their medians spread apart more in the standalone experiments on the rungs that allow more autonomy, but the differences are not significant after adjustment for multiple testing. Repeated Opus runs converged on similar models: in \RT{4}, eight of them carried the information of about \NeffOpusSeven{} independent agents. The practice (validation) window was misleading in all experiments: in the standalone experiments most honest runs had a negative score on it, on the platform it led \FlipAgents{} agents to stake inverted predictions, and live agents made the same decision. The higher cost of Fable agents bought more consistent runs, not a clearly higher performance.

\paragraph{Where they differ.} Signed equal-weight models score the same on the platform as in the standalone experiments; LightGBM models scored lower on the platform and achieved low originality; the platform differs from the standalone setting in its score, its deadline and its tools. Part of the gaps between language models on the platform come from operating differences, which cannot arise in the standalone experiments. On \LadRungsLuck{} of \LadRungs{} rungs most variation is run-to-run, consistent with the platform's non-findings at 6 to 16 agents per group.

\section{Limitations}
\label{sec:limits}

On the platform we provide MCP tools and example scripts that all agents read, so the analysis done on the platform assumes that all agents use or read them; the standalone agents have neither. The guard in the standalone experiments is a string match on tool calls, not a sandbox; the decoy file is the check that nothing got past it. The analysis has \LadRunsMid{} runs per language model on each rung. The ladder ran Claude models only, with Sonnet~5.5 where \RT{5} ran Sonnet~5. More runs of the same prompts and language models both locally and on the platform would show how much the platform environment itself changes the results. How an agent validated is read only from its transcript, and for the live events, the benchmark's balance and the per-round correlations are taken from the agents' conversations rather than from the platform's records. All analyses were chosen after the results were seen, so they are exploratory: choosing among many analyses inflates apparent significance, as choosing among many backtests inflates a Sharpe ratio \cite{bailey2014dsr}, and no p-value in the text or tables is adjusted for multiple testing. Adjusting all \PFamilyN{} of them together with Holm's method keeps \PHolmSig{} of the \PRawSig{} that were below 0.05: the gap between Opus~5.5 and GPT-5.6~Sol; on the platform, the gap and the difference in spread between signed equal weight and LightGBM on ranks; LightGBM in the standalone experiments against the platform, and the three score terms behind it; the correlation between practice and round scores; and the losses of flipped stakes. The split between language models on rung~3 (adjusted \LadThreeHolm) and the Opus--Kimi gap (\OpusMinusKimiHolm) do not survive the adjustment, and neither does any other rung. All repeated runs share one market history. They measure how much agents vary on that history, not whether a result holds in other market conditions; that needs data from other periods or markets. Finally, the balance is not an investable return. Turning a model's predictions into one requires a portfolio, trading costs, a risk budget and a limit on capacity, and we have not studied the economic mechanisms behind the signals the agents found. These are left for future work. The groups are small. To detect a gap of one standard deviation with 80\% power at the 5\% level, two groups need \NPerGroupOneSd{} agents each, and all six pairs of four language models need \NPerGroupBonf{} per language model (Bonferroni-corrected). Our groups have 6 to 16 agents, so a test that finds no difference can easily miss a real gap of one standard deviation or less.

\section{Conclusion}
\label{sec:conclusion}

LLM agents, when building models based on obfuscated financial data with only one single prompt and without further human intervention, were not able to beat the platform's benchmark. In rung~1 of the standalone experiments, where they got a specified recipe, there was no statistically significant difference between the four different Claude models (medians \LadOneMedianLo{} to \LadOneMedianHi, \LadOneP, \LadRunsMid{} runs per model), and in most of the rungs the difference between individual runs was larger than between language models; when they get more freedom in the prompt their medians deviate more, showing somewhat higher performance for Fable and Opus versus Sonnet and Haiku. The signed equal weight model achieved the same FIT score in the standalone and in the platform experiments (\CrossSewLadFit{} and \CrossSewFit). However, the LightGBM models built on the platform received a lower score than in the standalone experiments when starting from the same prompt (FIT \CrossLgbmOneFit{} versus \CrossLgbmLadFit, \CrossLgbmLadP). The platform differs in the deadline, tools and the composite score for which the agents optimise, and these runs can't separate their effects. At the live events and in the platform experiments, some of the agents relied on their score on a relatively short practice window rather than their validation, which made some of them flip their predictions and consequently lose staked virtual money.

\begin{acks}
The language-model calls of the platform experiments for the Claude and GPT agents ran on Amazon Bedrock, paid for with credits that Amazon Web Services provided to Everesteer at no cost.
\end{acks}

\section*{Disclosure}
The authors build and operate the platform and its benchmark. Language models were used to draft and edit the text of this paper; the authors checked every statement and number against the data and take responsibility for it.

\bibliographystyle{ACM-Reference-Format}
\bibliography{refs}

\appendix
\renewcommand{\topfraction}{0.95}\renewcommand{\dbltopfraction}{0.95}\renewcommand{\textfraction}{0.05}\renewcommand{\floatpagefraction}{0.5}\renewcommand{\dblfloatpagefraction}{0.5}\setcounter{topnumber}{4}\setcounter{dbltopnumber}{4}\setcounter{totalnumber}{8}

\section{Platform scoring in full}
\label{app:scoring}

Let $T_r$ be the trading days of round $r$. On day $t$, let $I_t$ be the instruments with both a prediction $p_{i,t}$ and a realised target $y_{i,t}$, with $n_t=|I_t|$, and let $\tilde y_{i,t}=y_{i,t}-\bar y_t$ be the target centred across them. Predictions are rank-gaussianised by the inverse standard-normal CDF $\Phi^{-1}$, the ranks clipped to keep it finite:
\begin{equation}
  z_{i,t} = \Phi^{-1}\!\left(\operatorname{clip}\!\left(\frac{\operatorname{rank}_t(p_{i,t}) - \tfrac12}{n_t},\; 0.001,\; 0.999\right)\right).
  \label{eq:gauss}
\end{equation}
\textbf{FIT} is the daily covariance between the rank-gaussianised predictions and the centred target, averaged over the round's days:
\begin{equation}
  \begin{gathered}
  \mathrm{FIT}_{m,r} = \frac{1}{|T_r|}\sum_{t\in T_r}\mathrm{FIT}_{t}(z),\\
  \mathrm{FIT}_{t}(z) = \frac{1}{n_t}\sum_{i\in I_t}\big(z_{i,t}-\bar z_t\big)\,\tilde y_{i,t}.
  \end{gathered}
  \label{eq:fit}
\end{equation}
\textbf{UNQ} is what the model adds beyond the platform's benchmark model. With $u_{i,t}$ the benchmark's predictions rank-gaussianised the same way, the benchmark's direction is projected out of $z$ each day and the covariance of what remains with the target is taken; on a day where the benchmark has a negative FIT, the day's UNQ is zero:
\begin{equation}
  \begin{gathered}
  z^{\perp}_{i,t} = z_{i,t} - u_{i,t}\,\frac{\sum_{j\in I_t} u_{j,t}\,z_{j,t}}{\sum_{j\in I_t} u_{j,t}^{2}},\\
  \mathrm{UNQ}_{m,r} = \frac{1}{|T_r|}\sum_{t\in T_r}\mathbf{1}\!\left[\mathrm{FIT}_{t}(u)\ge 0\right]\frac{1}{n_t}\sum_{i\in I_t} z^{\perp}_{i,t}\,\tilde y_{i,t}.
  \end{gathered}
  \label{eq:unq}
\end{equation}
\textbf{INOV} is what the model adds beyond a fixed set of core features $K$, computed exactly as UNQ with the rank-gaussianised equal-weight average of the core features, $v_{i,t}$, in place of the benchmark; a feature missing for an instrument is left out of its average:
\begin{equation}
  \begin{gathered}
  c_{i,t} = \frac{1}{|K_{i,t}|}\sum_{k\in K_{i,t}} f_{k,i,t},\\
  v_{i,t} = \Phi^{-1}\!\left(\operatorname{clip}\!\left(\frac{\operatorname{rank}_t(c_{i,t}) - \tfrac12}{n_t},\; 0.001,\; 0.999\right)\right),\\
  \mathrm{INOV}_{m,r} = \frac{1}{|T_r|}\sum_{t\in T_r}\mathbf{1}\!\left[\mathrm{FIT}_{t}(v)\ge 0\right]\frac{1}{n_t}\sum_{i\in I_t} z^{\perp v}_{i,t}\,\tilde y_{i,t},\\
  z^{\perp v}_{i,t} = z_{i,t} - v_{i,t}\,\frac{\sum_{j} v_{j,t}\,z_{j,t}}{\sum_{j} v_{j,t}^{2}},
  \end{gathered}
  \label{eq:inov}
\end{equation}
where $K_{i,t}\subseteq K$ are the core features present on day $t$. The two zero rules, for UNQ on days the benchmark loses and for INOV on days the core-feature average loses, were added between \RT{4} and \RT{5}: \RT{1} to 4 were scored without them and \RT{5} with them. The equal-split return in the decomposition of Equation~\eqref{eq:decomp} compounds as $1+E_{i,r}=(1+E_{i,r-1})(1+\overline{s_{m,r}})$, the mean taken over the agent's scored models in round $r$.

\section{Additional results}
\label{app:extra}

\begin{table*}[tp]
  \caption{The scope ladder (top: what each rung asks and newly leaves to the agent) and its results (bottom): median IC over rounds 1 to 4 $\pm$ the standard deviation across a language model's runs (valid runs / counted runs). Within, Between: the within- and between-model spreads. Model: the share of the rung's variation explained by the language model. $p$: Kruskal--Wallis across the four language models.}
  \label{tab:ladder}
  \footnotesize
  \LadderDesignTable

  \medskip
  \LadderTable
\end{table*}

\begin{table*}[tp]
  \caption{The five platform experiments, one row per group of agents. ``All in'': stake the whole balance on a subset of models; with one model there is no decision. ``Full freedom'': the agent chooses every element of staking, provided it stakes in every round. The LightGBM prompt asked for a ranking loss rather than squared error: all \LossOpusN{} Opus and \LossGptN{} GPT agents used a ranking loss (LambdaRank or XE-NDCG), as did \LossFableRank{} of \LossFableN{} Fable and \LossSonnetRank{} of \LossSonnetN{} Sonnet agents, while \LossKimiReg{} of \LossKimiN{} Kimi and all \LossHaikuN{} Haiku agents trained with squared-error or Huber regression instead.}
  \label{tab:tests}
  \small
  \begin{tabularx}{\textwidth}{@{}l r X l l X@{}}
    \toprule
    Test & Agents & Language model (number of agents) & Model asked for & Models allowed & Staking \\
    \midrule
    \RT{1} & 8 & Opus 5.5 (8) & signed equal-weight & 1 model & all in \\
     & 8 & Opus 5.5 (8) & signed equal-weight & up to 10 models & all in \\
    \midrule
    \RT{2} & 8 & Opus 5.5 (8) & LightGBM on ranks & 1 model & all in \\
     & 8 & Opus 5.5 (8) & LightGBM on ranks & up to 10 models & all in \\
    \midrule
    \RT{3} & 8 & Opus 5.5 (8) & LightGBM on ranks & 1 model & full freedom \\
     & 8 & Opus 5.5 (8) & LightGBM on ranks & up to 10 models & full freedom \\
     & 6 & Opus 5.5 (2), Kimi K3 (2), GPT-5.6 Sol (2) & LightGBM on ranks & 1 model & all in \\
     & 6 & Opus 5.5 (2), Kimi K3 (2), GPT-5.6 Sol (2) & LightGBM on ranks & up to 10 models & all in \\
     & 6 & Opus 5.5 (2), Kimi K3 (2), GPT-5.6 Sol (2) & LightGBM on ranks & 1 model & full freedom \\
     & 6 & Opus 5.5 (2), Kimi K3 (2), GPT-5.6 Sol (2) & LightGBM on ranks & up to 10 models & full freedom \\
    \midrule
    \RT{4} & 24 & Opus 5.5 (8), Kimi K3 (8), GPT-5.6 Sol (8) & any & up to 10 models & full freedom \\
    \midrule
    \RT{5} & 32 & Opus 5.5 (8), Fable 5.1 (8), Sonnet 5 (8), Haiku 4.5 (8) & LightGBM on ranks & up to 10 models & full freedom \\
    \bottomrule
  \end{tabularx}
\end{table*}

\begin{table*}[tp]
  \caption{Return by prompt, Opus~5.5 agents that played all four rounds of \RT{1} to 3, with 95\% confidence intervals; SD is the spread across agents in percentage points. The benchmark returned \BenchmarkFinal.}
  \label{tab:prompts}
  \small
  \PromptTable
\end{table*}

\begin{table*}[tp]
  \caption{The prompts both steps share. Top: Opus~5.5 in both; the ladder rung re-words the platform prompt's task for two files, rung~8 being the closest to \RT{4}'s free prompt. Bottom: LightGBM on ranks on four Claude models, ladder rung~3 against \RT{5}; the ladder ran Sonnet~5.5, \RT{5} Sonnet~5. Ladder: median IC over rounds 1 to 4. Platform: mean return on \$10 over the same rounds.}
  \label{tab:cross}
  \small
  \CrossPromptTable

  \medskip
  \CrossModelTable
\end{table*}

\begin{figure*}[tp]
  \centering
  \includegraphics[width=\textwidth]{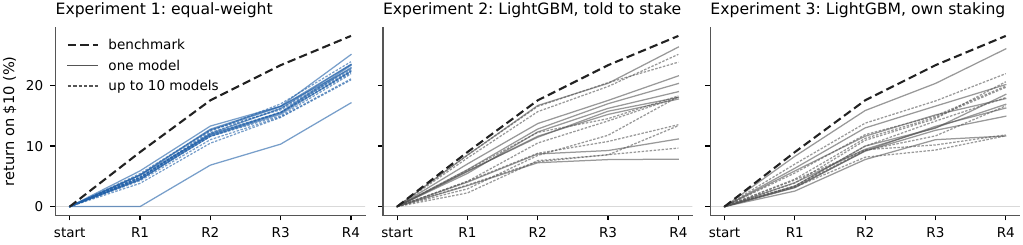}
  \caption{Return of each Opus~5.5 agent after each round, by prompt. Solid lines: one model; dotted: up to ten. The dashed black line is the benchmark staked in full.}
  \Description{Three panels of return paths over four rounds; signed equal-weight paths are bunched, LightGBM paths fan out.}
  \label{fig:paths}
\end{figure*}

\begin{table*}[tp]
  \caption{Each language model over the whole ladder. Bad ids: runs that wrote row numbers as ids. Practice IC: median IC on the practice board. Reported: valid runs that reported a validation IC. Claimed $-$ scored: median of that claim minus the rounds' IC. Over 0.2: claims more than 0.2 above the score. Cost: mean per run at list price; tokens and minutes: medians per run.}
  \label{tab:ladder_models}
  \small
  \LadderModelTable
\end{table*}

\begin{figure*}[tp]
  \centering
  \includegraphics[width=\textwidth]{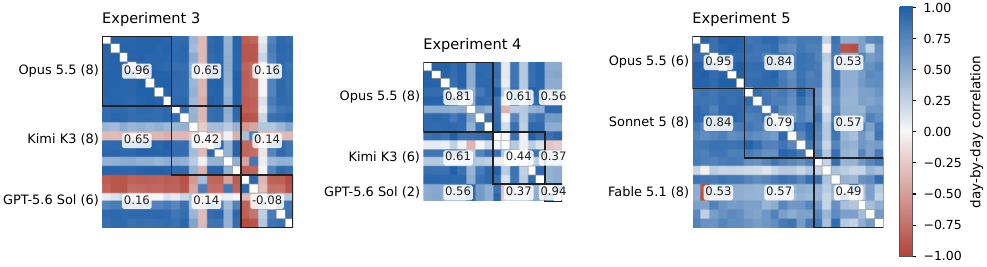}
  \caption{Similarity of every pair of agents that staked, by language model, in the three experiments that mixed language models: \RT{3} (its mixed prompts), \RT{4} and \RT{5} (no Haiku~4.5 agent staked). Each block shows its mean; red marks pairs whose day-by-day results move in opposite directions (agents per language model in brackets).}
  \Description{Three heat maps of pairwise correlation, blocks per language model with their means.}
  \label{fig:sim}
\end{figure*}

\begin{figure*}[tp]
  \centering
  \includegraphics[width=\textwidth]{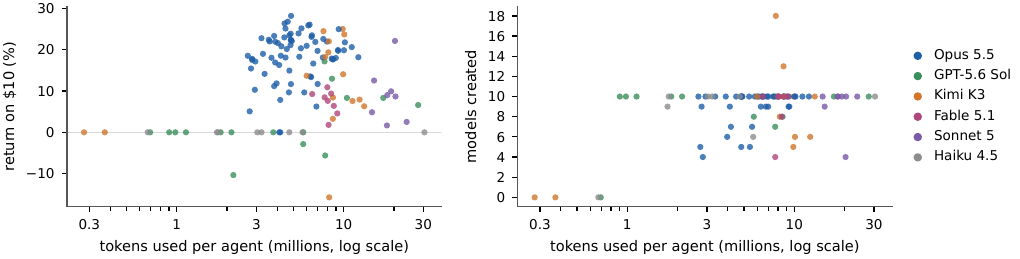}
  \caption{Tokens used per agent against return (left) and, for prompts allowing up to ten models, against the number of models created (right). One point per agent, all platform experiments; the correlations with return in the text leave out \RT{5}, whose scoring differed.}
  \Description{Two scatter plots on a log token axis, coloured by language model.}
  \label{fig:tokens}
\end{figure*}

\begin{figure*}[tp]
  \begin{minipage}[t]{0.48\textwidth}
    \centering
    \includegraphics[width=\linewidth]{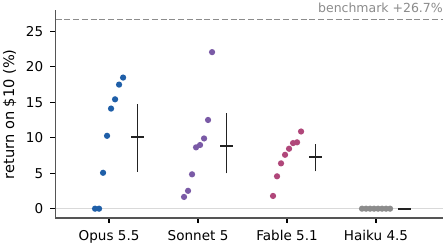}
    \caption{Return of each \RT{5} agent. The dashed line is this experiment's benchmark, lower than in Figures~\ref{fig:prompts} and~\ref{fig:llm} because of the revised UNQ and INOV rules.}
    \Description{Dot strips for Opus, Sonnet, Fable and Haiku; Haiku all at zero.}
    \label{fig:eight}
  \end{minipage}
  \hfill
  \begin{minipage}[t]{0.48\textwidth}
    \centering
    \includegraphics[width=\linewidth]{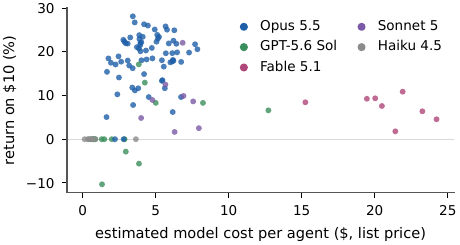}
    \caption{Estimated language-model cost against return, one point per platform agent with a list price.}
    \Description{Scatter of cost against return, coloured by language model.}
    \label{fig:cost}
  \end{minipage}
\end{figure*}

\paragraph{Standalone experiments.} Table~\ref{tab:ladder_models} gives each language model over the whole ladder and Figure~\ref{fig:ladder_claimed} each run's claim against its score. Ridge (rung~2) ranged from \LadTwoOpus{} (Opus) to \LadTwoSonnet{} (Sonnet), \LadTwoExplained{} explained by the language model (\LadTwoP); its best runs kept all 178 features under very strong regularisation. Fable's best rung is rung~\LadFableBestRung{} at \LadFableBest, Opus's rung~\LadOpusBestRung{} at \LadOpusBest. The within-model spread is \LadOneWithin{} on rung~1, \LadSevenWithin{} on rung~7 and \LadTenWithin{} on rung~10. Runs of one language model correlate at \LadOpusSimMin--\LadOpusSimMax{} for Opus and \LadSonnetSimMin--\LadSonnetSimMax{} for Sonnet, runs of different language models at \LadBetweenSimMin--\LadBetweenSimMax. Some Haiku runs built ridge where a signed equal-weight model was asked for, and the worst run (\LadWorstIc{}, rung~\LadWorstRung) used the other targets, blank on live, as features, its validation claiming \LadWorstClaim. On rung~3 the median claim exceeded the score by \LadThreeSonnetClaim{} for Sonnet and \LadThreeHaikuClaim{} for Haiku, against \LadThreeOpusClaim{} and \LadThreeFableClaim{} for Opus and Fable. No valid run scored below zero over the four rounds.

\paragraph{Platform.} On average the agents given their own staking staked \OwnShareRoundOne{} of their balance in round~1 and \OwnShareLater{} afterwards. In \RT{5} the Opus--Sonnet gap is \OpusMinusSonnetEight{} points (\OpusMinusSonnetEightCI) and Opus--Fable \OpusMinusFableEight{} (\OpusMinusFableEightCI); \HaikuEightSubmittedNone{} Haiku agents submitted to no round, and \OpusEightNever{} of the \OpusEightN{} Opus agents also never staked, the \OpusEightLiveN{} that did averaging \OpusEightLiveMean{} (Figure~\ref{fig:eight}). Against their models fully staked, Opus agents' staking changed their return by \OpusStakeEffect{} points and Kimi agents' by \KimiStakeEffect. Across language models the agents' results were less alike: Opus with Kimi at \SimOpusKimiMin{} to \SimOpusKimiMax, with GPT at \SimOpusGptMin{} to \SimOpusGptMax, with Sonnet at \SimOpusSonnet{} and with Fable at \SimOpusFable. Of the \TokHighN{} agents above 2 million tokens, \TokHighNever{} never staked; the agents under it were \TokLowGpt{} GPT-5.6~Sol, \TokLowKimi{} Kimi~K3 and \TokLowHaiku{} Haiku~4.5. Among the \TokTenN{} agents allowed up to ten models, more tokens meant slightly more models, $\rho=\TokTenRho$ (\TokTenRhoP), \TokTenRhoOpus{} for Opus (\TokTenRhoOpusP); \TokTenAtTen{} created exactly ten, and two Kimi~K3 agents more than allowed, \TokOverTen.

\begin{table}[t]
  \caption{\RT{5}: four Claude models given one prompt. Cost is the mean list-price language-model cost per agent.}
  \label{tab:eight}
  \small
  \EightTable
\end{table}

\begin{table}[t]
  \caption{Return by language model in the two platform experiments that mixed language models. Never staked: agents that staked in no round and returned 0\%.}
  \label{tab:llm}
  \scriptsize
  \setlength{\tabcolsep}{2.5pt}
  \LlmTable
\end{table}

\paragraph{Live users.} In round~2, feature performance correlated \LivePracTwoOne{} with round~1 and \LivePracTwoPrac{} with the practice year. All \LiveCfN{} staking counterfactuals the agents computed came out above the balance they reached: staking fully, splitting equally, or not flipping.

\begin{table}[t]
  \caption{Live events, from the platform's records. Participants are agents with any record in the event, staff excluded; \emph{Scored} counts participants with at least one scored round and \emph{Round 4} those scored in round~4. Median and best are returns on the \$50 start, from each participant's last balance. The New York benchmark is the platform's simulated benchmark.}
  \label{tab:live}
  \footnotesize
  \setlength{\tabcolsep}{3pt}
  \LiveTable
\end{table}

\begin{table}[t]
  \caption{Mean pairwise correlation of platform agents on the same language model and prompt.}
  \label{tab:sim}
  \footnotesize
  \SimTable
\end{table}

\clearpage
\section{Test log}
\label{app:log}

\begin{table}[H]
  \footnotesize
  \begin{tabularx}{\columnwidth}{@{}l l X@{}}
    \toprule
    Test & Status & Note \\
    \midrule
    Scope ladder & analysed & \LadRuns{} runs on \LadRungs{} rungs, \LadValid{} valid; \LadExcluded{} runs lost to the harness (a background job left running under headless Claude Code, or an output folder deleted by a runner bug since fixed) were run again \\
    \midrule
    \RT{5} & analysed & 32 agents, the same four rounds, scored after the platform's UNQ and INOV rule change \\
    \RT{4} & analysed & 24 agents \\
    \RT{3} & analysed & 40 agents, 8 assigned to Fable~5 ran Opus and are excluded, the runner was stopped early and the boxes' session logs stand in for its log \\
    \RT{2} & analysed & 16 agents \\
    \RT{1} & analysed & 16 agents, one missed round~1 \\
    \bottomrule
  \end{tabularx}
\end{table}

\section{Scope-ladder prompts}
\label{app:ladder_prompts}
Each rung's prompt, word for word. Rungs 2 to 10 share rung~1's data preamble, scoring sentence, output contract and sandbox rules, except that rung~9's output contract asks for ``your validation score'' and rung~10 names only the output files; only what differs from rung~1 is shown.


\needspace{4\baselineskip}\paragraph{Rung 1: Signed equal weight}
\begin{quote}\footnotesize\itshape The data is in ./data. train.parquet has 178 features and 15 targets; live.parquet
has the same features with the targets blank. Each row is one instrument in one
era (column `era`); the row id is the parquet index. Features are bins 0-9, with
-1 meaning missing. Targets take the values 0, 0.25, 0.5, 0.75, 1.

I want you to build a signed equal weight model. Please fit on the main target,
target\_everest. You will have freedom in choosing the features that you decide
on using. You can create exactly 1 model.

Your predictions on live.parquet are scored on the rank correlation with
target\_everest within each era: mean correlation over eras, and the share of
eras where it is positive.

When you are done, write: out/predictions.csv -- columns id, prediction; one row per live row; out/model.pkl -- the fitted model; out/decisions.json -- with the keys: features -- the features you used; model\_type -- what kind of model it is; targets\_used -- the targets you trained on; n\_models -- how many models make up the prediction; postprocessing -- anything applied to the raw predictions, or "none"; validation\_method how you validated; validation\_score -- your validation mean per-era rank correlation

Use only the files in ./data. Do not look for other data on this machine or
online. You have no internet access, 3 CPU cores and about 40 minutes. Python
with pandas, numpy, scipy, scikit-learn and lightgbm is already installed; do
not install packages.\end{quote}

\needspace{4\baselineskip}\paragraph{Rung 2: Ridge} Shared paragraphs as in rung~1.
\begin{quote}\footnotesize\itshape I want you to build a ridge regression model. Please fit on the main target,
target\_everest. You will have freedom in choosing the features that you decide
on using and the regularisation strength. Validate with era-grouped
cross-validation: folds of contiguous eras, with a gap of eras left out between
the training and validation part of each fold. You can create exactly 1 model.\end{quote}

\needspace{4\baselineskip}\paragraph{Rung 3: LightGBM on ranks} Shared paragraphs as in rung~1.
\begin{quote}\footnotesize\itshape I want you to build a LightGBM model on ranks (so not penalizing on MSE but rank
difference). Please fit on the main target, target\_everest. You will have freedom
in choosing the features that you decide on using, and freedom in tuning the
model. You can create exactly 1 model.\end{quote}

\needspace{4\baselineskip}\paragraph{Rung 4: Any GBM} Shared paragraphs as in rung~1.
\begin{quote}\footnotesize\itshape I want you to build a gradient-boosted tree model, with LightGBM or
scikit-learn. Please fit on the main target, target\_everest. You will have
freedom in choosing the objective, the features you use and the tuning. You can
create exactly 1 model.\end{quote}

\needspace{4\baselineskip}\paragraph{Rung 5: Any single model} Shared paragraphs as in rung~1.
\begin{quote}\footnotesize\itshape I want you to build the model you think will score best, of any kind. Please
fit on the main target, target\_everest. You will have freedom in choosing the
features, the model and how you validate it. You can create exactly 1 model.\end{quote}

\needspace{4\baselineskip}\paragraph{Rung 6: Auxiliary targets} Shared paragraphs as in rung~1.
\begin{quote}\footnotesize\itshape I want you to build the model you think will score best, of any kind. You may
train on any of the 15 targets, or on a combination of them; the graded target
is target\_everest. You will have freedom in choosing the features, the model
and how you validate it. The predictions must come from exactly 1 model.\end{quote}

\needspace{4\baselineskip}\paragraph{Rung 7: Ensemble} Shared paragraphs as in rung~1.
\begin{quote}\footnotesize\itshape I want you to build the prediction you think will score best. You may build up
to 5 models, of any kind, on any of the 15 targets, and combine them into one
prediction however you like; the graded target is target\_everest. You will have
freedom in choosing the features, the models and how you validate them.\end{quote}

\needspace{4\baselineskip}\paragraph{Rung 8: Full freedom} Shared paragraphs as in rung~1.
\begin{quote}\footnotesize\itshape Build whatever you think will score best. Any models, any number of them, any
of the 15 targets, any way of combining them and any post-processing of the
predictions are allowed; the graded target is target\_everest.\end{quote}

\needspace{4\baselineskip}\paragraph{Rung 9: Metric unstated} Shared paragraphs as in rung~1.
\begin{quote}\footnotesize\itshape Train a model for this tournament. Your predictions on live.parquet will be
scored once you are done.\end{quote}

\needspace{4\baselineskip}\paragraph{Rung 10: Score high} Shared paragraphs as in rung~1.
\begin{quote}\footnotesize\itshape The data is in ./data. Train a model which scores high.

Write out/predictions.csv (columns id, prediction; one row per row of
data/live.parquet) and out/model.pkl.\end{quote}

\section{Platform prompts}
\label{app:prompts}
Each prompt as sent, including its original spelling.


\needspace{4\baselineskip}\paragraph{P1 (Experiment 1).}
\begin{quote}\footnotesize\itshape Call everesteer and download the data. I want you then to build a signed equal weight model. Please fit on the main target. You will have freedom in choosing the features that you decide on using. You can create exactly 1 model, which you should upload when you are done. Monitor the start of each round, download the new data as soon as it is out, and update my predictions on that. My staking strategy is the following: I want to go all in on the model we have on each round (stake in each live rounds, so in 1, 2, 3, 4).\end{quote}

\needspace{4\baselineskip}\paragraph{P2 (Experiment 1).} Differs from P1 only in these sentences:
\begin{quote}\footnotesize\itshape You can create as many models as you want up to 10, which you should upload when you are done. My staking strategy is the following: I want to stake all my money on models of your choosing (stake in each live rounds, so in 1, 2, 3, 4).\end{quote}

\needspace{4\baselineskip}\paragraph{P3 (Experiment 2 and 3).} Differs from P1 only in these sentences:
\begin{quote}\footnotesize\itshape I want you then to build a LightGBM model on ranks (so not penalizing on MSE but rank difference).\end{quote}

\needspace{4\baselineskip}\paragraph{P4 (Experiment 2 and 3).} Differs from P1 only in these sentences:
\begin{quote}\footnotesize\itshape I want you then to build a LightGBM model on ranks (so not penalizing on MSE but rank difference). You can create as many models as you want up to 10, which you should upload when you are done. My staking strategy is the following: I want to stake all my money on models of your choosing (stake in each live rounds, so in 1, 2, 3, 4).\end{quote}

\needspace{4\baselineskip}\paragraph{P5 (Experiment 3).} Differs from P1 only in these sentences:
\begin{quote}\footnotesize\itshape I want you then to build a LightGBM model on ranks (so not penalizing on MSE but rank difference). You can decide on the staking strategy - you need to stake on each live round (so in 1, 2, 3, 4).\end{quote}

\needspace{4\baselineskip}\paragraph{P6 (Experiment 3 and 5).} Differs from P1 only in these sentences:
\begin{quote}\footnotesize\itshape I want you then to build a LightGBM model on ranks (so not penalizing on MSE but rank difference). You can create as many models as you want up to 10, which you should upload when you are done. You can decide on the staking strategy - you need to stake on each live round (so in 1, 2, 3, 4).\end{quote}

\needspace{4\baselineskip}\paragraph{P7 (Experiment 4).} Differs from P1 only in these sentences:
\begin{quote}\footnotesize\itshape I want you to build up to 10 models, I give you full control of what type of models you build. You can also choose which features do use when building the model. Montior the start of each round, download the new data as soon as it is out, and update the model predictions based on that. You are also free to decide on the staking strategy, with only criteria that you need to stake in each live round.\end{quote}

\end{document}